%% file: main.tex
\documentclass[10pt,twocolumn,letterpaper]{article}

\usepackage[final]{cvpr}              
\usepackage{graphicx}
\usepackage{multirow}
\usepackage{tcolorbox}
\usepackage{xcolor}
\usepackage{listings}
\usepackage{fancyvrb}

\definecolor{cvprblue}{rgb}{0.21,0.49,0.74}
\usepackage[pagebackref,breaklinks,colorlinks,allcolors=cvprblue]{hyperref}

\def\paperID{*****} 
\def\confName{CVPR}
\def\confYear{2023}
\def\approach{LaDeCo}

\newtcolorbox{coloredtextbox}[2][]{
    colback=#2!10,             
    colframe=#2,               
    coltitle=white,            
    colbacktitle=#2,           
    title=#1,                  
    fonttitle=\bfseries,       
    sharp corners=south,       
    arc=2mm,                   
    boxrule=1mm,               
    width=0.5\textwidth,          
}

\newtcolorbox{coloredtextbox1}[2][]{
    colback=#2!10,             
    colframe=#2,               
    coltitle=white,            
    colbacktitle=#2,           
    title=#1,                  
    fonttitle=\bfseries,       
    sharp corners=south,       
    arc=2mm,                   
    boxrule=1mm,               
    width=\textwidth,          
}

\title{From Elements to Design: A Layered Approach for Automatic Graphic \\ Design Composition}

\author{
    Jiawei Lin\textsuperscript{1},
    Shizhao Sun\textsuperscript{2},
    Danqing Huang\textsuperscript{2},
    Ting Liu\textsuperscript{1},
    Ji Li\textsuperscript{2},
    Jiang Bian\textsuperscript{2} \\
    \textsuperscript{1}Xi’an Jiaotong University, \textsuperscript{2}Microsoft Research \\
    \texttt{\small kylelin@stu.xjtu.edu.cn, tingliu@mail.xjtu.edu.cn,} \\
    \texttt{\small \{shizsu, dahua, jili5, jiabia\}@microsoft.com}
}

\begin{document}
\twocolumn[{%
    \renewcommand\twocolumn[1][]{#1}%
    \maketitle
    \begin{center}
        \centering
        \includegraphics[width=\textwidth]{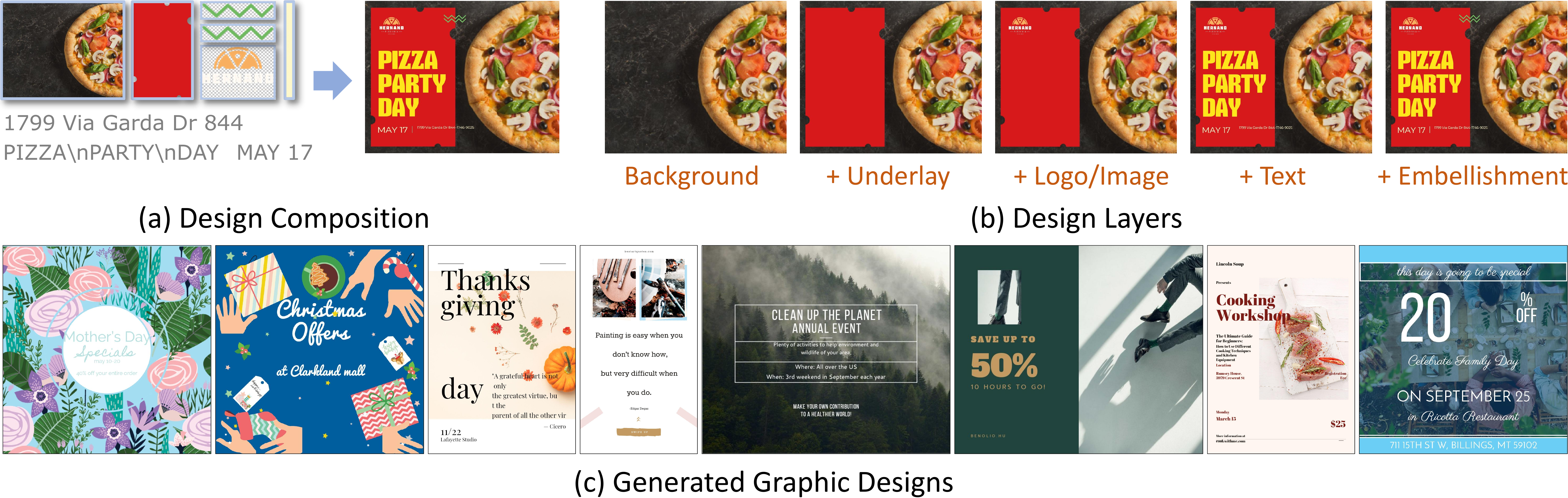}
        \captionof{figure}{(a) Given a set of multimodal elements as input, our approach automatically composes them into a cohesive, balanced, and aesthetically pleasing graphic design. (b) Since a holistic design can be divided into different layers according to element semantics, we achieve the design composition task in a layer-by-layer manner. (c) Our approach is able to craft high-quality design pieces.}
        \label{fig: teaser}
    \end{center}%
}]

\maketitle

\input{sec/0_abstract}    
\input{sec/1_intro}
\input{sec/2_related_work}

\input{sec/3_method}

\input{sec/4_exp}

\input{main.bbl}

\input{sec/X_suppl}

\end{document}

%% file: sec/0_abstract.tex
\begin{abstract}

In this work, we investigate automatic design composition from multimodal graphic elements.
Although recent studies have developed various generative models for graphic design, they usually face the following limitations: they only focus on certain subtasks and are far from achieving the design composition task; they do not consider the hierarchical information of graphic designs during the generation process.
To tackle these issues, we introduce the layered design principle into Large Multimodal Models (LMMs) and propose a novel approach, called \approach{}, to accomplish this challenging task.
Specifically, \approach{} first performs layer planning for a given element set, dividing the input elements into different semantic layers according to their contents.
Based on the planning results, it subsequently predicts element attributes that control the design composition in a layer-wise manner, and includes the rendered image of previously generated layers into the context.
With this insightful design, \approach{} decomposes the difficult task into smaller manageable steps, making the generation process smoother and clearer.
The experimental results demonstrate the effectiveness of \approach{} in design composition.
Furthermore, we show that \approach{} enables some interesting applications in graphic design, such as resolution adjustment, element filling, design variation, etc.
In addition, it even outperforms the specialized models in some design subtasks without any task-specific training.
\end{abstract}

%% file: sec/1_intro.tex
\section{Introduction}
\label{sec:intro}

Graphic design is an artistic discipline dedicated to creating visual content that attracts attention and communicates messages effectively.
Creating visually appealing designs today relies on human designers with both artistic creativity and technical expertise to skillfully integrate multimodal graphic elements like images, headlines, and decorative embellishments, etc.
This is a complex and time-consuming process that requires careful consideration of many aspects.
For example, as shown in Figure~\ref{fig: teaser}a, it is important to ensure that the main object (i.e., pizza) is not obscured by other elements.
For readability, there should be sufficient contrast between the text and the underlay.
Additionally, designers also need to adjust the element sizes to make the design balanced.
In this work, we refer to the challenging process of composing a set of elements into a holistic design as \emph{design composition} (see Figure~\ref{fig: teaser}a).

To ease the burden on human designers, recently, there has been a growing interest in developing generative models to streamline this process.
Most existing work focuses on certain typical subtasks of design composition.
For example, some previous approaches investigate content-aware layout generation~\cite{seol2024posterllama, yang2024posterllava, zhou2022composition, horita2024retrieval, hsu2023posterlayout}, which aims to automatically arrange graphic elements on a given canvas while ensuring that the main object remains unobstructed.
Although these methods are capable of creating high-quality layouts, they typically only consider the background image content while overlooking the content of other elements.
In addition, they do not predict text-related attributes during the layout generation process, limiting their ability to produce fully integrated designs.
Another popular subtask is called typography generation~\cite{inoue2023towards, gao2023textpainter, shimoda2024towards, peong2024typographic, jia2023cole}.
Its goal is to generate font, color, size, and other attributes for text elements, enhancing both aesthetics and readability.
However, it ignores the visual elements in graphic designs.
Together, all these studies fall short of holistic design creation.
Consequently, users have to manually integrate models of different functions to achieve design composition, which brings high costs and unnecessary obstacles.

To be best of our knowledge, FlexDM~\cite{inoue2023towards} is the only attempt towards automatic design composition.
By representing graphic designs as a flat combination of element attributes, FlexDM formulates various design tasks as masked field prediction problems, including the design composition task.
While the flat representation provides practicality and versatility, it has a notable limitation: it overlooks the inherent hierarchical structure within graphic designs.
This hierarchical structure arises because human designers usually follow the layered design principle.
Specifically, it involves arranging elements in separate semantic layers, starting with a background and then gradually adding underlays, images, texts and small embellishments (see Figure~\ref{fig: teaser}b).
Such layered principle brings two main benefits. 
First, each preceding layer provides a strong foundation for designing subsequent layers, aiding cohesion. 
Second, grouping similar elements by layer clarifies the design process and enhances workflow efficiency.

\begin{figure*}[t]
    \centering
    \includegraphics[width=\linewidth]{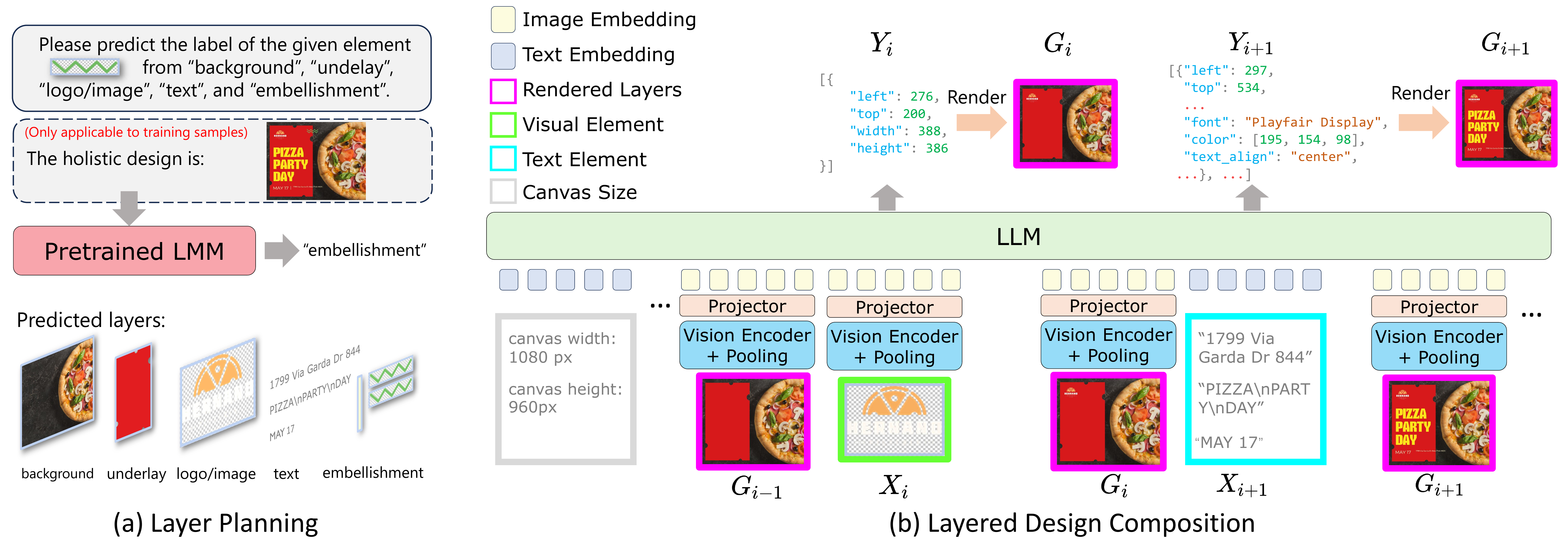}
    \caption{Illustration of our proposed \approach{}. First, it utilizes GPT-4o~\cite{gpt4o} to annotate the semantic labels for input elements. The layer structure is obtained from the predictions. Then \approach{} fine-tunes LMMs to achieve layered design composition. After generating each layer, the intermediate designs will be rendered as images and fed back into LMMs to guide subsequent layer generation.} 
    \vspace{3mm}
    \label{fig:approach}
\end{figure*}

Based on the insight, we propose \textit{\approach{}} (see Figure~\ref{fig:approach}), a layered design composition method built upon Large Multimodal Models (LMMs)~\cite{achiam2023gpt, zhu2023minigpt, team2023gemini, liu2024visual, chen2024internvl}.
Considering that input elements are inherently multimodal and that design composition can be formulated by sequentially predicting attributes for each element, 
we choose LMMs as the backbone for this task.
To support the layered mechanism, we develop a layer planning module.
Specifically, we carefully design the task prompts and leverage GPT-4o~\cite{gpt4o} to predict the semantic label for each input element by considering its content.
Elements sharing the same label are placed on the same layer, thus reaching layer planning.
Subsequently, \approach{} divides the generation process into several steps according to the layer planning results.
At each step, LMMs are asked to predict element attributes within a single layer.
After each step, the previously generated layers will be rendered as an intermediate design image and fed back into the LMMs, providing contextual information for following layer predictions.

Thanks to the novel design, \approach{} decomposes the challenging task into smaller manageable steps.
At each step, the model only focuses on design composition of the current layer, making the process more accessible.
Additionally, by rendering intermediate designs and adding them into the context, the model can better generate subsequent layers based on previous layers.
The layerwise generation also offers flexibility to allow \approach{} supporting certain subtasks of design composition without any task-specific training, as shown in Section~\ref{sec:task-specific}.

To validate the effectiveness of our approach, we conduct experiments on a publicly available dataset Crello~\cite{yamaguchi2021canvasvae} to compare it with the state-of-the-art baselines.
Specifically, the comparisons cover the full design composition task and its subtasks: content-aware layout generation and typography generation.
Both quantitative and qualitative experimental results show that \approach{} significantly outperforms the baseline models in design composition.
Through ablation studies, we demonstrate the effectiveness of the layer planning module and layered design composition introduced in \approach{}.
Besides, we show that \approach{} enables some interesting applications such as resolution adjustment, element filling, and design variation, further showcasing the practicality and versatility of \approach{}.
In addition, \approach{} even surpasses the specialized models in the two subtasks.


%% file: sec/2_related_work.tex
\section{Related Work}
\label{sec:related_work}

\paragraph{Graphic Design Composition.}

There are three aspects of relevant research on graphic design composition: content-aware layout generation, typography generation, and the holistic design composition.
We'll introduce them below.

\textit{Content-Aware Layout Generation}~\cite{hsu2023posterlayout, seol2024posterllama, yang2024posterllava, horita2024retrieval, lin2024layoutprompter, cheng2024graphic}.
It is an emerging research topic that studies layout generation conditioned on a given canvas.
LayoutPrompter~\cite{lin2024layoutprompter} proposes a RAG-based approach to retrieve training samples with the most similar saliency bounding boxes as prompts, and achieves content-aware layout generation via in-context learning.
PosterLlama~\cite{seol2024posterllama} introduces a LLM-based two-stage training method for this task.
It first keeps the backbone parameters fixed and trains the adapter, and then fine-tunes the backbone to generate layouts in the HTML format.
PosterLLaVA~\cite{yang2024posterllava} further improves practicality by supporting more user requirements (e.g., element relationships).
Compared to our work, these existing methods do not consider element content, nor predict text attributes, and thus cannot create a complete design.

\textit{Typography Generation}~\cite{gao2023textpainter, peong2024typographic, inoue2023towards, inoue2024opencole, jia2023cole}.
This task studies visually compelling, harmonious text rendering on a graphic design.
Some relevant work~\cite{gao2023textpainter, peong2024typographic} regards text styles as images and formulates typography generation as text image generation in the pixel space.
For editability, some other work directly predicts editable text attributes (e.g., font type, color, size) and utilizes the renderer to visualize stylized text elements.
FlexDM~\cite{inoue2023towards} achieves this by masked field prediction, generating all attributes in a single pass.
COLEs~\cite{jia2023cole, inoue2024opencole} propose Typography-LMM to autoregressively predict text attributes based on the input canvas.
However, these methods still can not craft holistic graphic designs since they do not consider the visual elements.

\textit{Design Composition}.
FlexDM~\cite{inoue2023towards} can accomplish design composition by simultaneously masking element positions and text attributes.
However, it treats all input elements equally and does not incorporate the critical layered information during the generation process.
In summary, we are the first to introduce layered design composition and develop a novel approach based on it.

\vspace{-5mm}
\paragraph{Other Topics in Graphic Design.}

Earlier, some studies investigate \textit{content-agnostic layout generation}.
To meet diverse user needs, previous methods have considered various input conditions, such as element attributes~\cite{kong2022blt, kikuchi2021constrained, jiang2023layoutformer++, inoue2023layout, li2020attribute, lin2024layoutprompter, zhang2023layoutdiffusion}, element relationships~\cite{kikuchi2021constrained, lee2020neural}, textual descriptions~\cite{lin2023parse, lin2024layoutprompter}, and more.
However, the content-agnostic nature makes them unable to adapt to the input elements, in which element deformations often occur.
Recently, there is an interesting topic studying \textit{design content generation}~\cite{inoue2024opencole, jia2023cole, kikuchi2024multimodal}.
We emphasize that we focus on design composition from user-provided elements rather than generate the element content as presented in their work.

\vspace{-5mm}
\paragraph{Large Multimodal Models (LMMs).}

LMMs have shown strong capabilities in understanding multimodal contexts and generating plausible responses across diverse domains~\cite{liu2024visual, maaz2023video, ye2023mplug, gpt4o, zhu2023minigpt, achiam2023gpt}.
In this work, we leverage them to effectively achieve design composition from multimodal input elements.
We introduce the layered design principle in our approach, enabling LMMs to tackle different semantic layers iteratively instead of generating all layers at once.
To the best of our knowledge, we are the first work to adopt LMMs for layered design generation.
Furthermore, our layerwise generation method also reflects the idea of chain-of thought (CoT) reasoning~\cite{wei2022chain}, which has been proven to be effective in enhancing reasoning performance~\cite{zhang2023multimodal, mitra2024compositional}.


%% file: sec/3_method.tex
\section{\approach{}}
\label{sec:method}

\subsection{Problem Formulation}
\label{sec:formulation}

The input and output of the design composition task are a set of multimodal elements and the element attributes, respectively.
When obtaining the predicted attributes, we can adopt an off-the-shelf renderer to render the input elements into a holistic graphic design, thereby achieving design composition.
In this work, input elements are categorized into two modalities: image modality and text modality.
For image modality elements, the output attributes are four bounding box parameters, i.e., left and top coordinates, element width and height.
For text modality elements, we consider eight more attributes, namely angle, font, font size, color, text alignment, capitalization, letter spacing, and line height.
We empirically find that these attributes are sufficient to describe a high-quality design.

\subsection{Method Overview}
\label{sec:overview}

In this work, we take inspiration from the layered design principle to decompose a holistic graphic design into different layers and progressively create these layers to reach the complete design, making the design composition process smoother and clearer.
Here, a layer is a collection of graphic elements with same semantic labels.
To be more specific, our method includes two key techniques, namely a layer planning module and a layered design composition process.
The layer planning module is responsible for categorizing input elements into pre-defined layers.
Then, in the layered design composition process, our approach predicts element attributes in a layerwise manner and gathers them together to get the complete attributes.

\subsection{Layer Planning}
\label{sec:layer_planning}

The very first step here is to determine a reasonable layer structure.
By examining numerous completed design pieces and consulting experienced designers, in this work, we consider 5 design layers, namely background, underlay, logo/image, text, and embellishment layers in the placement order (see Figure~\ref{fig: teaser}b).
By sequentially rendering these layers on the empty canvas $G_0$, we obtain $G_1$ through $G_5$ (see Figure~\ref{fig:layer}), where $G_1$ represents only the background layer, $G_2$ includes the background layer plus the underlay layer, and so forth, with $G_5$ representing the complete, finalized design.
Notably, the layer structure is not limited to our proposed one.
There is flexibility to add or remove some as long as it is reasonable.

Although the publicly available dataset does not contain any layer information for the elements, we find it is feasible to infer from element content.
For example, a solid colored rectangular box might be an underlay, and a element with a star is an embellishment.
Therefore, we formulate layer planning as an element content understanding problem and leverage pre-trained LMMs to resolve it.
In the implementation, we employ GPT-4o~\cite{gpt4o} to automatically generate input element labels (see Figure~\ref{fig:approach}a), thereby achieving layer planning.
To guide GPT-4o effectively, we carefully craft prompts, clearly defining the problem, describing the characteristics of each element label, and requesting the model to output the most appropriate one for input elements.
For training samples where ground truth designs are available, we also include the design images and some metadata (e.g., the canvas size, element size) into the model input to enhance the prediction accuracy.
For full details of the prompt text, please refer to the supplementary materials.

\subsection{Layered Design Composition}
\label{sec:layered_composition}

Our main idea here is to generate element attributes in a layerwise manner from background to embellishment layers, according to the layer planning results in Section~\ref{sec:layer_planning}.
The rendered images of previous layers are incorporated in the context for the prediction of the subsequent layers.
It is noteworthy that this process requires strong understanding capabilities of element content.
For example, as shown in the row 3, column 1 of Figure~\ref{fig:qualitative}, the text element with an email address should be placed at the bottom of the canvas, and the barbershop logo should be placed at the top.
Meanwhile, the understanding of intermediate results is also critical.
For example, in the same figure, the model should avoid blocking the persons by understanding the intermediate rendered image.
To this end, we resort to LMMs, which possess remarkable understanding capabilities, to model the layered design process for multimodal design elements. 
As shown in Figure~\ref{fig:approach}b, in the $i$-th design layer, LMMs predict the attribute set $Y_i$ for current layer's elements $X_i$.
The preceding layers are rendered as an image $G_{i-1}$ and reintroduced into the LMMs to be part of the contextual input, guiding the generation of $Y_i$.
Next, we will detail the representation of $X_i$ and $Y_i$, as well as the model architecture.

\noindent \textbf{Representation of $X_i$ and $Y_i$.}
In general, we represent $X_i$ and $Y_i$ by concatenating the element content and attributes of current layer, respectively.
For background, underlay, logo/image, and embellishment layers, the inputs are all visual elements.
Hence, $X_i$ is a combination of element images (i.e., pixel values).
For the text layer, we concatenate its text content together to construct $X_i$ (see Figure~\ref{fig:approach}b).
In terms of $Y_i$, we follow the trend of structured representation in existing approaches~\cite{inoue2024opencole, seol2024posterllama, lin2024layoutprompter}, and serialize the element attributes into JSON strings, as implemented in OpenCOLE~\cite{inoue2024opencole}.
Notably, within each layer, the elements are randomly shuffled to prevent information leakage (e.g., the top-down placement order).
In the design layers that do not have any elements, we represent their inputs as \texttt{null} and outputs as an empty JSON string \texttt{\{\}}.
Please refer to the supplementary materials for an example of the input-output representation.

\begin{table*}[th]
    \centering
    \begin{small}
    \begin{tabular}{lcccccccccc}
    \toprule
        \multirow{2}{*}{Methods} & \multicolumn{5}{c}{LLaVA-OV Scores } & \multirow{2}{*}{Val } & \multirow{2}{*}{Ove } & \multirow{2}{*}{Ali } & \multirow{2}{*}{Und$_l$ } & \multirow{2}{*}{Und$_s$ } \\
        & (i) & (ii) & (iii) & (iv) & (v) & & & & &  \\
        \midrule
        FlexDM~\cite{inoue2023towards} & 5.34 & 5.29 & 5.41 & 5.09 & 4.54 & 0.8757 & 0.3242 & \textbf{0.0016} & 0.7286 & 0.7298 \\
        GPT-4o~\cite{gpt4o} & 6.53 & 6.49 & 6.60 & 6.27 & 5.69 & 0.9968 & 0.0595 & 0.0001 & 0.3780 & 0.5708 \\
        \approach{} (Ours) & \textbf{8.08} & \textbf{7.92} & \textbf{8.00} & \textbf{7.82} & \textbf{6.98} &  \textbf{0.9365} & \textbf{0.0865} & 0.0013 & \textbf{0.6922} & \textbf{0.6580} \\
        \midrule
        GT & 8.35 & 8.21 & 8.30 & 8.01 & 7.26 & 0.9265 & 0.0768 & 0.0015 & 0.6848 & 0.6732 \\
        \bottomrule
    \end{tabular}
    \end{small}
    \caption{Quantitative comparison on the design composition task. LLaVA-OV evaluation includes the following aspects: (i) design and layout, (ii) content relevance, (iii) typography and color, (vi) graphics and images, and (v) innovation and originality. The score closest to the one calculated from real data (denoted as GT) is highlighted in bold, indicating the best performance among different methods.}
    \label{tab:quantitative_main}
\end{table*}

\noindent \textbf{Model Architecture.}
The model consists of three components: a vision encoder, a projector and the LMM backbone.
The vision encoder is responsible for encoding the element images and intermediate designs, generating image embeddings.
The projector then projects these embeddings to match the hidden state dimension required by the backbone.
Finally, the backbone is used to model the joint distribution across layers, ensuring cohesion in the layered design process.
To reduce computational complexity, a 2D average pooling operation is applied to the output of vision encoder to compress the image tokens effectively.

\subsection{Training and Inference}
\label{sec:training}


\noindent \textbf{Training.}
Ground truth attributes for training samples are available, allowing us to pre-render and cache the intermediate canvas states $\{G_i\}_{i=1}^5$ based on the layer planning results (see Figure~\ref{fig:layer}).
During training, we fine-tune the model by minimizing the negative log-likelihood of $Y_i$ across all layers:
\begin{equation}
\nonumber
    \mathcal{L} = -\sum_{i=1}^5 \log P(Y_i \vert Y_{<i}, X_{\leq i}, G_{< i}).
\end{equation}

\noindent \textbf{Inference.}
At inference time, \approach{} iteratively generates design layers from $G_1$ to $G_5$, thereby achieving the goal of design composition.
Compared to generating the whole attributes at once, \approach{} adds only about a $20\%$ increase in rendering time to obtain the intermediate designs, making it an effective and efficient method.

Notably, \approach{} offers remarkable flexibility at inference.
It can handle other design subtasks without any task-specific training. 
For example, when the ground truth background layer ($G_1$) is provided, and the model is tasked with generating $G_2$ through $G_5$, it can effectively achieve content-aware layout generation.
Similarly, when $G_1$ to $G_3$ are given as ground truth, and the model is asked to generate $G_4$, it performs typography generation.

%% file: sec/4_exp.tex
\section{Experiments}
\label{sec:exp}

\subsection{Setup}
\noindent\textbf{Datasets.}
We conduct experiments on the publicly available Crello-v4~\cite{yamaguchi2021canvasvae} dataset, which includes 23,421 graphic designs sourced from VistaCreate~\footnote{\href{https://create.vista.com/}{https://create.vista.com/}}.
Since Crello has provided separate rendered pixel images for all elements, we can conveniently build the training inputs as described in Section~\ref{sec:layered_composition}.
Besides, based on the pixel images as well as the layer planning results obtained in Section~\ref{sec:layer_planning}, we can easily render the intermediate designs for training samples through the renderer~\footnote{\href{https://github.com/CyberAgentAILab/OpenCOLE/tree/main/src/opencole/renderer}{https://github.com/CyberAgentAILab/OpenCOLE/src/opencole/renderer}} developed in OpenCOLE~\cite{inoue2024opencole}.
We adopt the same data splits of Crello-v4, dividing the dataset into 19,095 training, 1,951 validation, and 2,375 test samples.
To enhance training efficiency, we filter out training samples with more than 25 elements (a total of 938 examples). 
In addition, we gather a large-scale commercial dataset similar to Crello to study the effect of dataset size on performance.
We refer to it as LargeCrello.
LargeCrello dataset contains a total of 109,235 samples.
We also filter out its samples exceeding 25 elements and manually verify that LargeCrello has no overlap with the test set of Crello.

\noindent\textbf{Implementation Details.}
We choose Llama-3.1-8B~\footnote{\href{https://huggingface.co/meta-llama/Llama-3.1-8B}{https://huggingface.co/meta-llama/Llama-3.1-8B}}, one of the most advanced open-source large language models (LLMs), as the model backbone.
The vision encoder is initialized from the CLIP ViT-L/14 model~\footnote{\href{https://huggingface.co/openai/clip-vit-large-patch14-336}{https://huggingface.co/openai/clip-vit-large-patch14-336}}, and the projector is structured as a two-layer MLP using GELU~\cite{hendrycks2016gaussian} activation functions.
To enable efficient training, we utilize the LoRA technique~\cite{hu2021lora} on the backbone, jointly optimizing LoRA parameters and the projector while keeping the vision encoder parameters fixed.
We conduct training on four A100-80G GPUs with a global batch size of 128, and employ AdamW~\cite{loshchilov2017decoupled} to optimize for about 7K iterations with a learning rate of 2e-4.
As for the hyper-parameters, the rank number of LoRA is set to 32, the rank alpha is 64, and the token number of an input image is set to 5 (1 \texttt{cls} token, $2 \times 2$ compressed tokens).
At inference, we set the sampling temperature to 0.7 and Top-p (nucleus sampling) to 0.95, balancing diversity and quality for generated designs.

\begin{figure*}[t]
    \centering
    \includegraphics[width=1.02\linewidth]{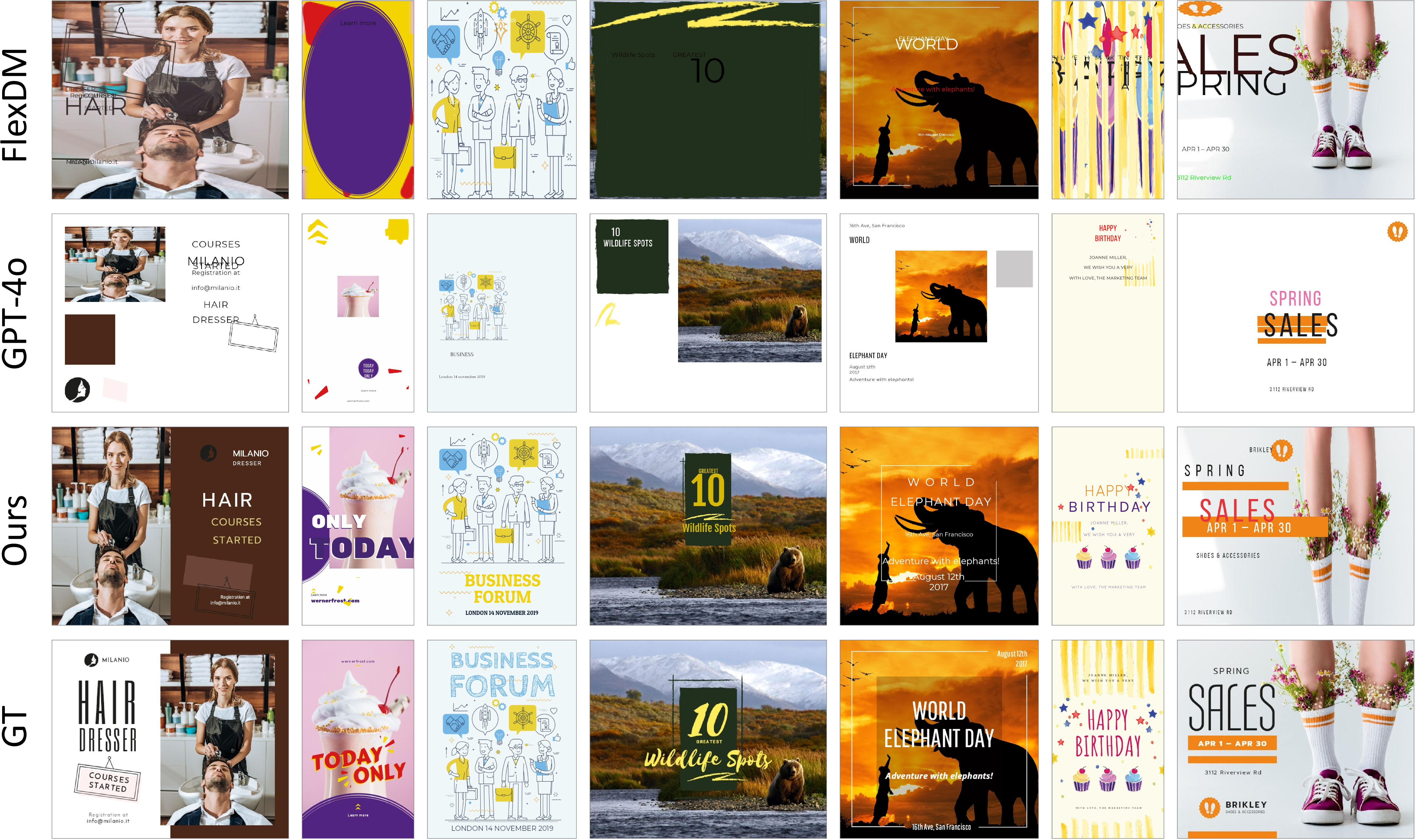}
    \caption{Qualitative comparison. We also show the ground truth designs for these samples. Please zoom in for a better view.}
    \label{fig:qualitative}
\end{figure*}

\begin{figure}
    \centering
    \includegraphics[width=\linewidth]{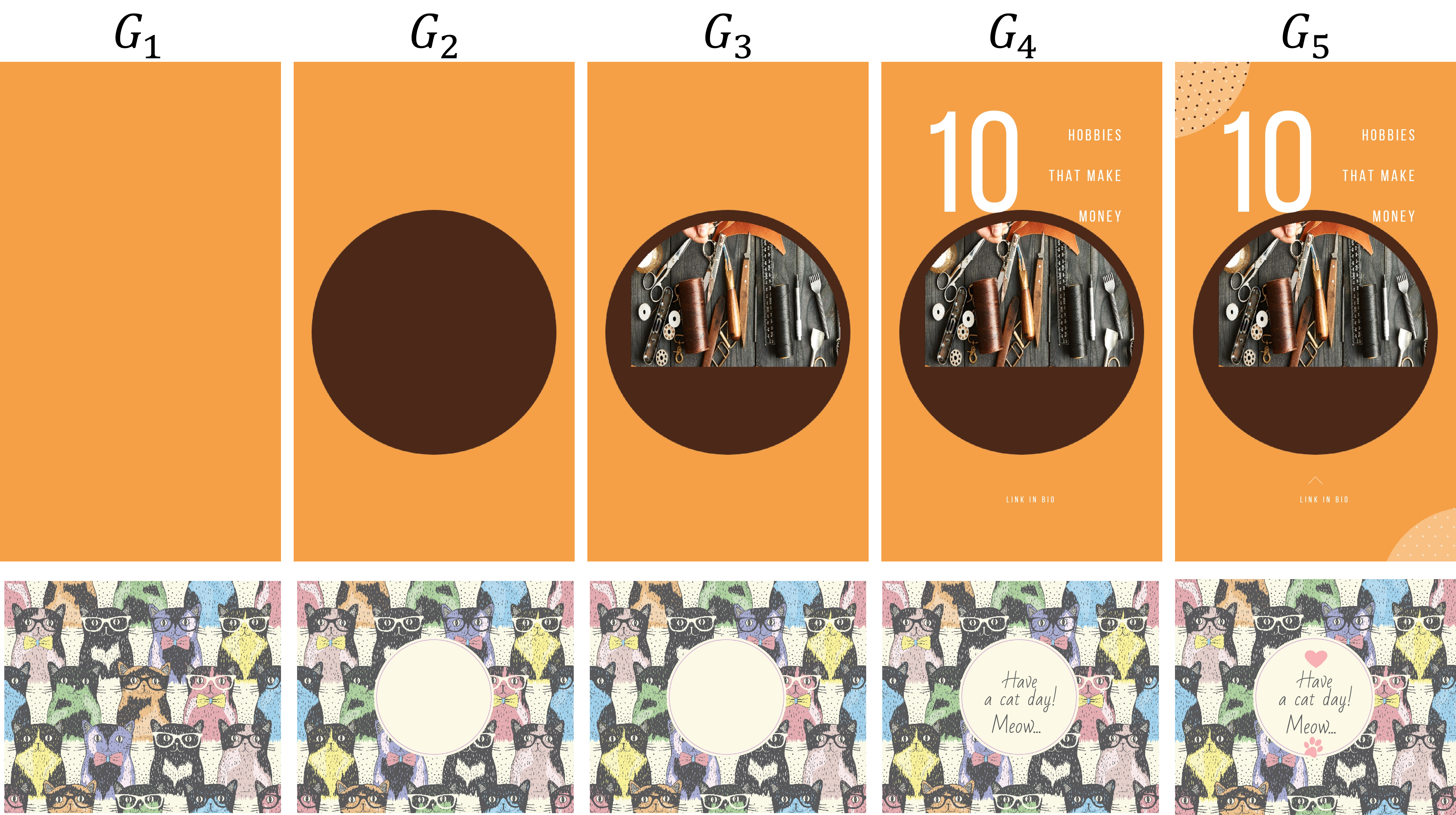}
    \caption{The rendered results of different layers from \approach{}.}
    \label{fig:layer}
\end{figure}

\noindent\textbf{Baselines.}
To demonstrate the effectiveness of \approach{}, we compare it with existing methods, i.e., FlexDM~\cite{inoue2023towards} and GPT-4o~\cite{gpt4o}.
FlexDM originally conducts experiments on Crello-v2, which has different dataset splits of Crello-v4.
We re-train it on the upgraded v4 dataset.
To accommodate design composition, we mask the position and text attribute fields, and predict them simultaneously.
In the GPT-4o baseline, we sequentially concatenate element contents in a manner similar to our approach, and prompt GPT-4o to generate their attributes one by one.
For the categorical attributes (e.g., font), we provide all options in the context.

\noindent\textbf{Evaluation Metrics.}
We follow previous work to prepare the evaluation metrics.
(1) \emph{Overall metrics.}
Following COLEs~\cite{jia2023cole, inoue2024opencole}, we introduce a robust proxy model for comprehensive evaluation.
Specifically, we use the LLaVA-OV-7B model~\cite{li2024llava} to evaluate quality across five aspects: design and layout, content relevance, typography and color, graphics and images, innovation and originality.
We use the same prompts as presented in COLE~\cite{jia2023cole}.
(2) \emph{Geometry-related metrics.}
These metrics focus purely on the geometric attributes of elements without considering their content, including element validity (Val), overlap (Ove), alignment (Ali), and underlay effectiveness (Und$_l$, Und$_s$)~\cite{hsu2023posterlayout, seol2024posterllama, lin2024layoutprompter}.
For each metric, the closer it is to the one calculated from real data (denoted as GT in the table), the better.
Note that higher or lower values alone are not indicative of better performance.
For example, a model could always put all the elements along the left side of the canvas to achieve a low alignment (Ali) score, but such an arrangement would not produce meaningful and high-quality designs.

\begin{table*}[t]
    \centering
    \begin{small}
    \begin{tabular}{lcccccccccc}
    \toprule
        \multirow{2}{*}{Settings} & \multicolumn{5}{c}{LLaVA-OV Scores } & \multirow{2}{*}{Val } & \multirow{2}{*}{Ove } & \multirow{2}{*}{Ali } & \multirow{2}{*}{Und$_l$ } & \multirow{2}{*}{Und$_s$ } \\
        & (i) & (ii) & (iii) & (iv) & (v) & & & & & \\
        \midrule
        Llama-3.1-8B (rank 16) & 8.03 & 7.89 & 8.00 & 7.75 & 6.90 & 0.9347 & 0.0796 & 0.0012 & 0.6900 & 0.6564 \\
        Llama-3.1-8B (rank 64) & 8.10 & 7.94 & 8.04 & 7.83 & 6.98 & 0.9352 & 0.0787 & 0.0013 & 0.7084 & 0.6715 \\
        \midrule
        llava-v1.5-7b (rank 32) & 8.00 & 7.86 & 8.02 & 7.78 & 6.90 & 0.9403 & 0.0940 & 0.0015 & 0.6703 & 0.6208 \\
        Llama-3.1-8B-Instruct (rank 32) & 8.08 & 7.89 & 8.03 & 7.82 & 6.99 & 0.9388 & 0.0804 & 0.0015 & 0.6867 & 0.6640  \\
        \midrule
        w/o LP, w/o LDC (rank 32) & 7.23 & 7.12 & 7.28 & 6.99 & 6.29 & 0.9325 & 0.0954 & 0.0013 & 0.6194 & 0.5875 \\
        w/ LP, w/o LDC (rank 32) & 7.84 & 7.67 & 7.78 & 7.56 & 6.66 & 0.9389 & 0.0843 & 0.0013 & 0.6568 & 0.6242 \\
        \midrule
        Llama-3.1-8B* (rank 32) & 8.22 & 8.06 & 8.22 & 7.94 & 7.09 & 0.9335 & 0.1029 & 0.0005 & 0.7321 & 0.7116 \\
        \midrule
        Llama-3.1-8B (rank 32) & 8.08 & 7.92 & 8.00 & 7.82 & 6.98 & 0.9365 & 0.0865 & 0.0013 & 0.6922 & 0.6580 \\
        \midrule
        GT & 8.35 & 8.21 & 8.30 & 8.01 & 7.26 & 0.9265 & 0.0768 & 0.0015 & 0.6848 & 0.6732 \\
        \bottomrule
    \end{tabular}
    \end{small}
    \caption{Ablation studies. Our investigation covers four aspects (from top to bottom): (1) the rank number in LoRA, (2) the base model, (3) the key techniques in \approach{}, where \textit{LP} denotes layer planning , and \textit{LDC} represents layered design composition, (4) dataset size.
    The model with * to is trained on the combined Crello and LargeCrello datasets, while the models without * are trained on Crello only.}
    \vspace{-2mm}
    \label{tab:ablation_study}
\end{table*}

\subsection{Quantitative Evaluation}
Table~\ref{tab:quantitative_main} shows quantitative results.
\approach{} significantly outperforms the baseline models in overall metrics (i.e., LLaVA-OV scores), indicating the superiority of \approach{} in design composition.
Particularly, \approach{} achieves very good scores in \textit{design and layout} as well as \textit{typography and color}. 
This suggests that \approach{} excels at layout generation and nuanced attribute prediction, both of which are critical for design creation.
In terms of geometry-related metrics, \approach{} achieves closet scores to the one calculated on real data (denoted as GT) on most metrics, showcasing its strong capabilities to model real-world data. 
In contrast, baseline models often struggle with some metrics. 
For example, FlexDM exhibits a serious overlap issue, while GPT-4o has low underlay effectiveness.

\begin{figure}
    \centering
    \includegraphics[width=\linewidth]{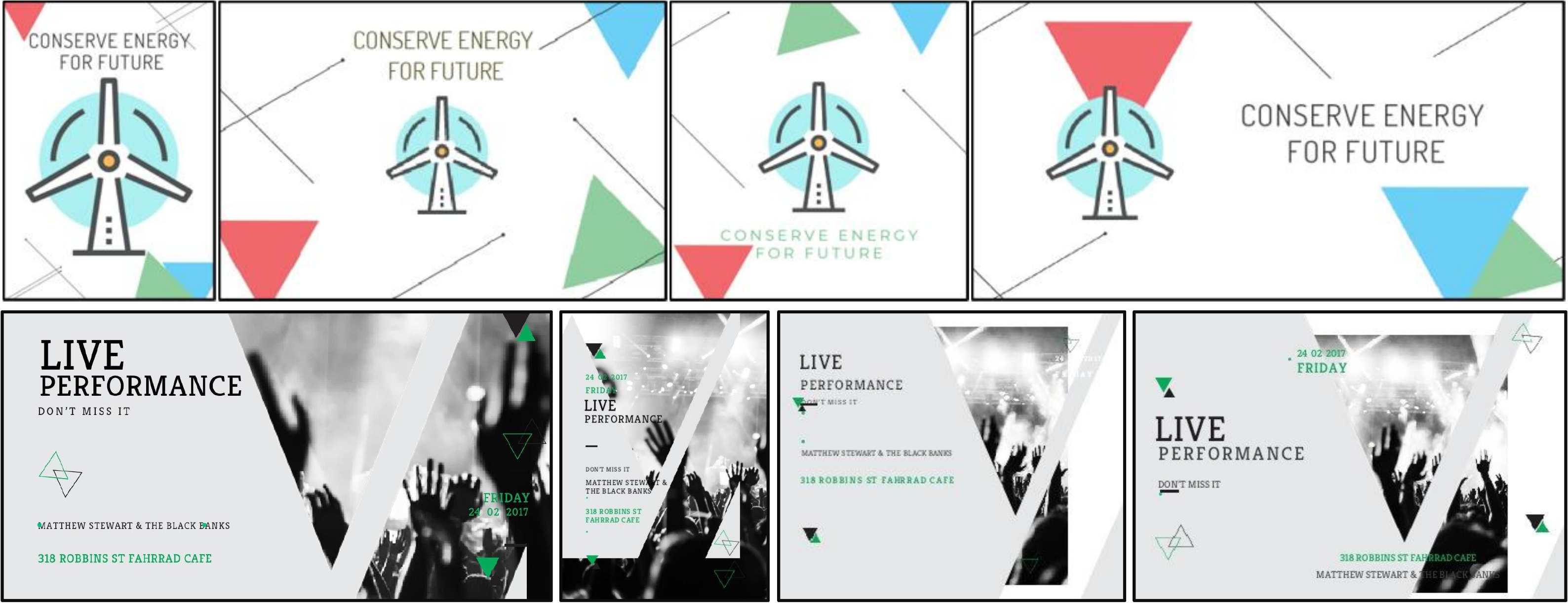}
    \vspace{-2mm}
    \caption{\approach{} composes the same input elements to designs with different canvas sizes.}
    \vspace{-2mm}
    \label{fig:different_ar}
\end{figure}

\begin{figure}
    \centering
    \includegraphics[width=\linewidth]{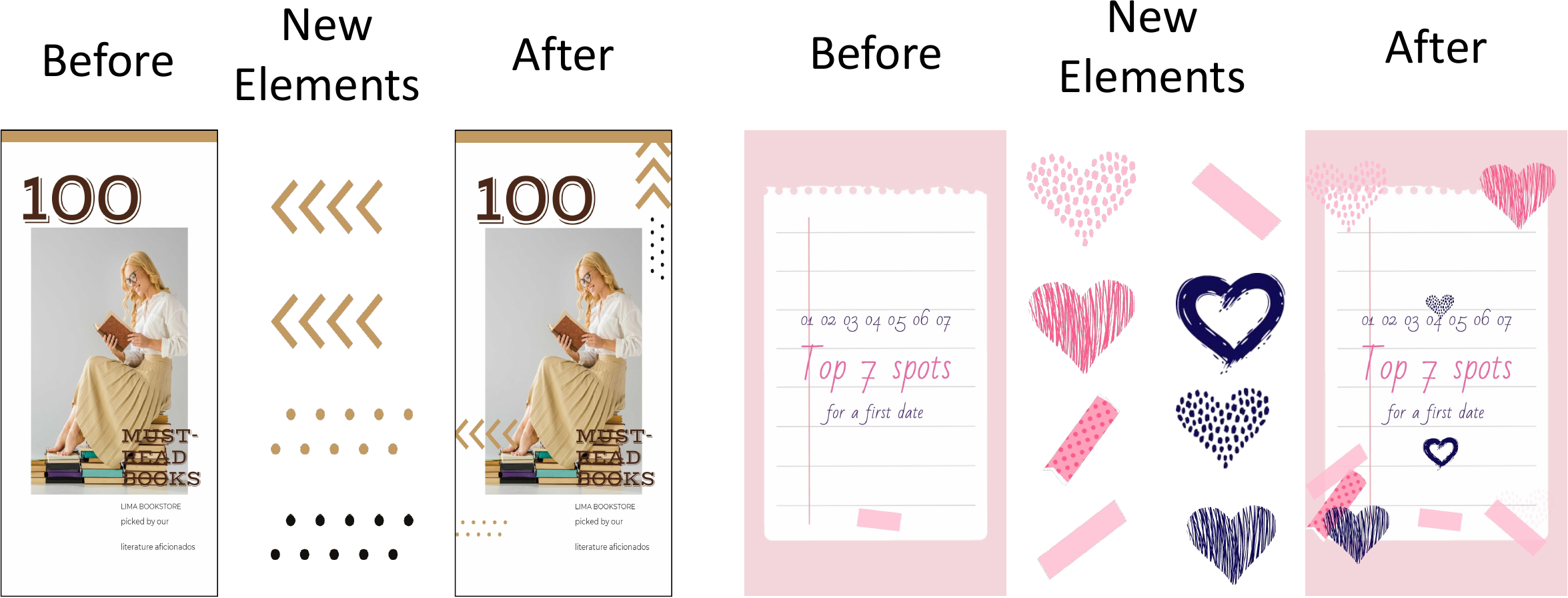}
    \vspace{-2mm}
    \caption{\approach{} adds new elements on a existing design to achieve a more appealing design.}
    \vspace{-2mm}
    \label{fig:element_filling}
\end{figure}

\subsection{Qualitative Evaluation}

We show rendered generated designs for each method and the corresponding ground truth designs in Figure~\ref{fig:qualitative}.
The results demonstrate that \approach{} is proficient in composing input elements into high-quality, visually pleasing and balanced designs.
On the contrast, the baselines all suffer from some serious problems, such as composition failure (FlexDM, column 2, 6), poor readability (FlexDM, column 1) and imbalance (GPT-4o, column 1, 3, 4).
Additionally, \approach{} can accurately capture relationships between elements.
For instance, text elements are precisely positioned on underlay elements (Ours, column 1, 4, 5, 7), and embellishment elements contribute meaningfully to decorate the designs (Ours, column 2, 6).

We further shows the rendered results of different layers generated by \approach{} in Figure~\ref{fig:layer}.
These results demonstrate that with the proposed layer planning module, \approach{} can generate meaningful layers.

\begin{figure}
    \centering
    \includegraphics[width=\linewidth]{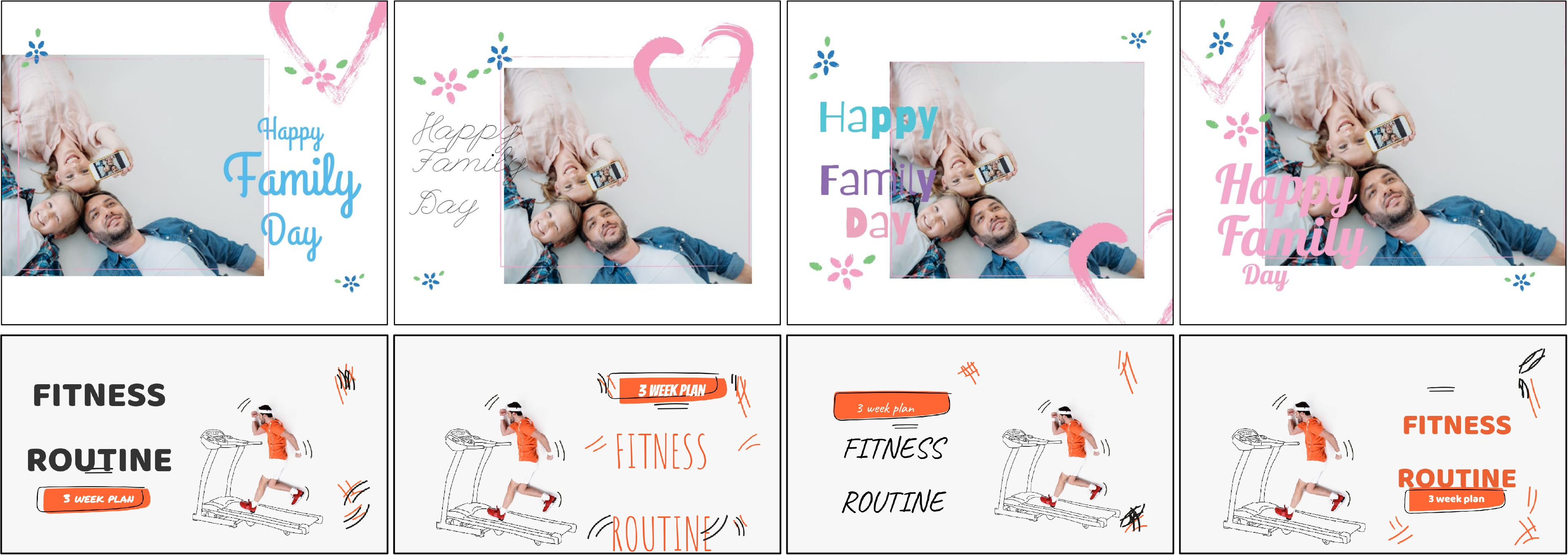}
    \vspace{-2mm}
    \caption{\approach{} creates diverse designs with the same elements.}
    \vspace{-2mm}
    \label{fig:diversity}
\end{figure}


Besides, \approach{} also enables some interesting applications in graphic design. 
Figure~\ref{fig:different_ar} shows that \approach{} can achieve design composition on the condition of different canvas sizes (called \emph{resolution adjustment}).
The predicted attributes will be adjusted to suit the canvas size, making the final designs appealing in various canvas sizes.
Figure~\ref{fig:element_filling} presents that \approach{} can add new element on a existing design to make it more pleasing (called \emph{element filling}).
Figure~\ref{fig:diversity} shows that the given the same input elements, \approach{} can compose them to create diverse designs, which provides multiple choices to users (called \emph{design variations}).

\begin{table*}[t]
    \centering
    \begin{small}
    \begin{tabular}{lcccccccc}
    \toprule
        Methods & Val  & Ove  & Ali  & Und$_l$  & Und$_s$  & Uti  & Occ  & Rea \\
        \midrule
        PosterLLaVa~\cite{yang2024posterllava} & \textbf{0.9269} & 0.0685 & 0.0011 & 0.7879 & 0.7375 & 0.4199 & 0.1936 & \textbf{0.0747} \\
        PosterLlama~\cite{seol2024posterllama} & 0.8701 & 0.0868  & 0.0014 & 0.8483 & 0.7798 & 0.4115 & \textbf{0.1772} & 0.0694 \\
        \approach{} (Ours) & 0.9340 & \textbf{0.0805} & \textbf{0.0016} & \textbf{0.6851} & \textbf{0.6540} & \textbf{0.4414} & 0.1835 & 0.0768 \\
        \midrule
        GT & 0.9265 & 0.0768 & 0.0015 & 0.6848 & 0.6732 & 0.4737 & 0.1628 & 0.0709 \\
        \bottomrule
    \end{tabular}
    \end{small}
    \caption{Quantitative results on the content-aware layout generation subtask. The score closest to
the one calculated from real data (denoted as GT) is highlighted in bold, indicating the best performance among different methods.}
    \vspace{-3mm}
    \label{tab:quantitative_content_aware}
\end{table*}

\subsection{Ablation Studies}
Table~\ref{tab:ablation_study} shows the results of ablation studies.
(1) \textit{Rank number in LoRA}:
When the rank number varies from 16 to 32, and then to 64, the quantitative metrics show little variation, indicating the robustness of \approach{} with respect to the amount of training parameters.
(2) \textit{Base model}:
We adopt two other pre-trained LLMs, llava-v1.5-7b and Llama-3.1-8B-Instruct, as the backbone.
Similar to previous ablation study, our \approach{} is robust to the choice of base model.
(3) \textit{Key techniques in \approach{}}: 
We consider two settings.
First, we remove the layered design composition (LDC) and use the model to predict all element attributes sequentially without taking the  intermediate rendered layers as input (denoted as w/ LP, w/o LDC (rank 32) in the table).
This leads to drops in five overall metrics and two underlay effectiveness metrics, indicating the importance of LDC in design composition.
Second, we continue to remove layer planning (LP) module and construct the input-output representation using a random ordered design elements (denoted as w/o LP, w/o LDC (rank 32) in the table).
The aforementioned metrics show a significant decline, which indicates that arranging elements following the hierarchical structure does play a critical role in design composition.
(4) \textit{Dataset size}:
When the model is trained on the combined datasets, most quantitative metrics are considerably improved and get closer to ground truth values.
The results suggest that \approach{} is a scalable algorithm with respect to dataset size.

\begin{figure}
    \centering
    \includegraphics[width=\linewidth]{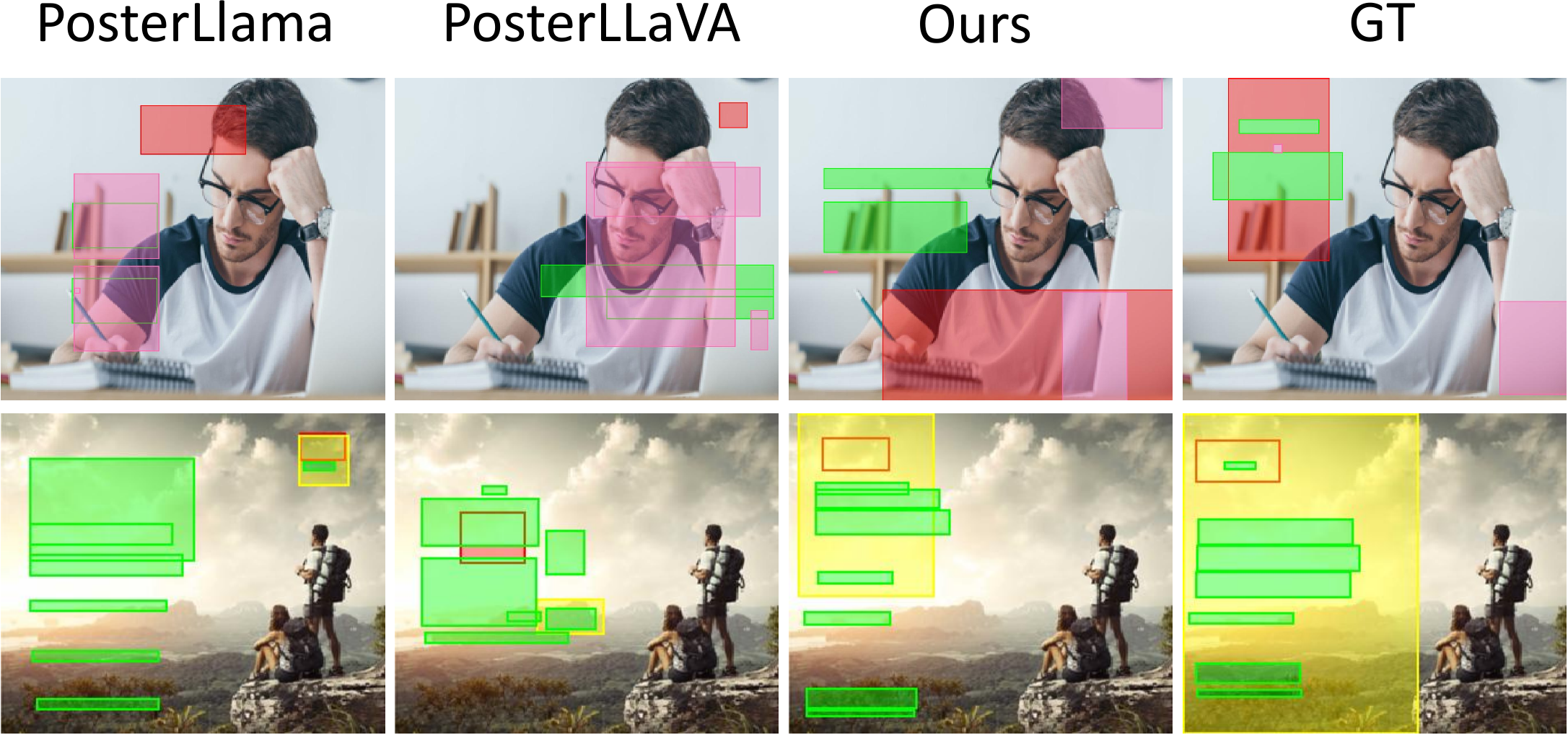}
    \caption{Qualitative comparison on the content-aware layout generation task. The yellow, red, green, pink boxes represent underlay, image, text, and embellishment elements, respectively.}
    \vspace{-2mm}
    \label{fig:qualitative_contentaware}
\end{figure}

\subsection{Comparison with Task-specific Baselines}
\label{sec:task-specific}
We further compare \approach{} with task-specific baselines, which focus on handling sub-tasks of design composition.

\emph{Content-aware layout generation}.
We consider two state-of-the-art baselines specialized for this subtask, including  PosterLLava~\cite{yang2024posterllava} and PosterLlama~\cite{seol2024posterllama}.
We re-train them on the Crello dataset for a fair comparison.
Besides geometry-related metrics, we further adopt metrics in DS-GAN~\cite{hsu2023posterlayout} to assess content-related quality based on the given canvas, i.e., canvas utility (Uti), occlusion (Occ), and text readability (Rea). 
Table~\ref{tab:quantitative_content_aware} shows quantitative results.
Compared to specialized methods, \approach{} achieves best performance on most metrics.
Figure~\ref{fig:qualitative_contentaware} shows qualitative results.
\approach{} excels in generating layouts that prevent blocking the main content within the given canvas.

\begin{figure}
    \centering
    \includegraphics[width=\linewidth]{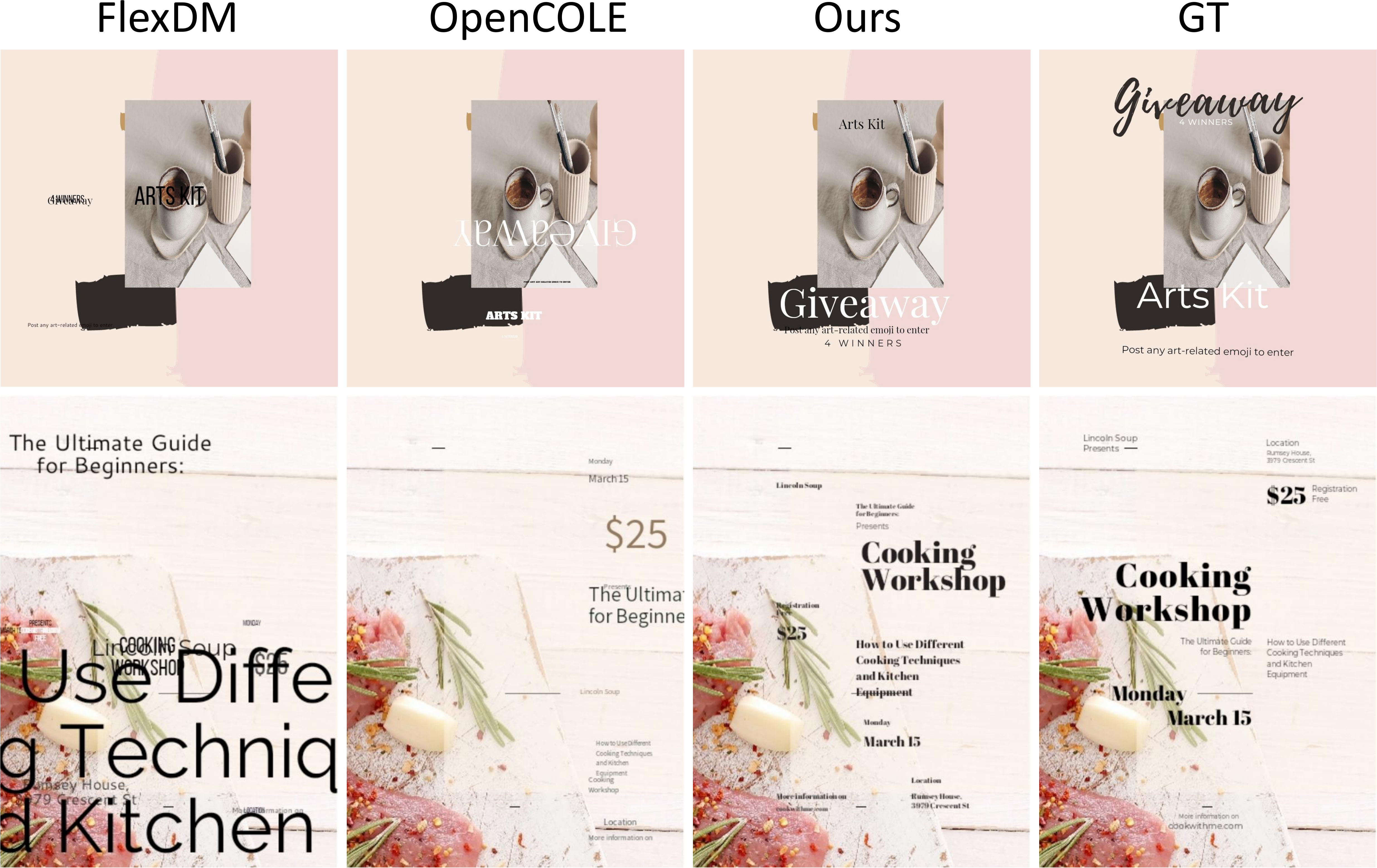}
    \caption{Qualitative comparison on typography generation. }
    \vspace{-2mm}
    \label{fig:qualitative_typography}
\end{figure}

\emph{Typography generation.} 
We consider FlexDM~\cite{inoue2023towards} and OpenCOLE~\cite{inoue2024opencole} as baselines.
For FlexDM, we only mask the text attribute fields, leaving the position fields accessible.
For OpenCOLE, which has already been trained on Crello-v4, we leverage the released Typography LMM model~\footnote{\href{https://huggingface.co/cyberagent/opencole-typographylmm-llava-v1.5-7b-lora}{https://huggingface.co/cyberagent/opencole-typographylmm}} for comparison.
Figure~\ref{fig:qualitative_typography} shows the resutls.
\approach{} takes into consideration various aspects such as text layout, aesthetics, and readability during typography generation process.
In contrast, the baseline models suffer from text overlap and poor readability.

\section{Conclusion}
In this work, we introduce \approach{} for design composition. 
The main idea is integrating the inherent hierarchical structure, which emerges from the layered principle during the practical design process, into LMMs. 
Specifically, a layer planning module categorizes input elements into different layers based on their contents, and a layered design composition process predicts the attributes that controls the composition by using the rendered previous layers as context.
In the future, we plan to expand our model to accommodate more user requirements, e.g., textual descriptions. 
We also plan to explore the integration of our design composition model with image generation models for element content creation, aiming to achieve end-to-end design generation.

\newpage

%% file: sec/X_suppl.tex
\clearpage
\maketitlesupplementary
\renewcommand\thesection{\Alph{section}}
\pagestyle{empty}
\setcounter{section}{0}

\section{Implementation Details of Layer Planning}

Layer planning is achieved by predicting semantic labels, including background, underlay, logo/image, text, and embellishment, for input elements.
In the implementation, we can easily identify text elements since they have the text content attribute.
For non-text elements, we then leverage GPT-4o~\cite{gpt4o} to categorize them into the other four labels in a zero-shot manner.
We carefully design the following \textit{Element Labeling Prompt} to accomplish this effectively.

\begin{coloredtextbox}[Element Labeling Prompt]{brown}
You are an excellent graphic designer.
Your task is to determine the role of the given element, which is rendered as an image.
There are 4 possible options: \textbf{Background}, \textbf{Underlay}, \textbf{Logo/Image} or \textbf{Embellishment}.
Please refer to the detailed descriptions below to make your prediction. 

\textbf{Background}: The foundational layer of the design, typically large in size and covering the entire canvas. It may consist of a solid color, gradient, landscape image, or similar visual foundation.

\textbf{Underlay}: A supportive layer placed beneath key content, often used to create contrast or highlight the main design elements, such as borders, buttons, color overlays, and so on.

\textbf{Logo/Image}: A core visual element that represents a brand, product, or entity. It combines both imagery and logo elements to capture attention and convey the primary message.

\textbf{Embellishment}: Decorative elements that enhance visual appeal without conveying core information. These elements add style to the design. Note that they are usually small in size.

When you respond, please output only one word from the 4 options.
Do not include any additional explanations or irrelevant information.

The element is \texttt{<image>}. Please predict the given element role: 
\end{coloredtextbox}

For training samples, we further include the complete design, canvas size, and element size in the prompt.
Such information helps enhance prediction accuracy. 
For instance, if an element is small in size, it is likely to be an embellishment.

\begin{coloredtextbox}[Additional Prompt for Training Samples]{purple}
The overall design is \texttt{<image>}. 
The canvas width is $w_\text{canvas}$px, canvas height is $h_\text{canvas}$px.
The element is \texttt{<image>}. 
The element width is $w_\text{element}$px, element height is $h_\text{element}$px. 
Please also consider the provided canvas and element width/height, as they might be helpful in making a decision.
Please predict the given element role: 
\end{coloredtextbox}

\section{Input-Output Structure in Layered Design Composition}

Here we present an example to demonstrate the model input and output in layered design composition.
The \texttt{<image>} token is a placeholder which is replaced by image embeddings during data loading, as described in Section~\ref{sec:layered_composition}.
For text elements, we directly put their content in the model input (e.g., \textit{Spring Clean}, \textit{Best hacks}).
Since all intermediate designs can be rendered in advance, the training process is parallelized across the five design layers, with the loss function only calculated in the \textbf{ASSISTANT} parts.
This ensures the training efficiency of \approach{}.

\section{More Qualitative Results}

We show more qualitative results and the layered design composition process in Figure~10 and Figure~11, respectively.
The results indicate that \approach{} is proficient in creating high-quality and visually appealing designs with the novel layered design composition technique.

\section{Interesting Applications in Graphic Design}

\approach{} enables interesting applications in graphic design, including design variation, resolution adjustment, and element filling.
We show more results in Figure~12, Figure~13, and Figure~14, respectively.

\section{More Qualitative Comparison}

Figure~15, 16, 17 show more qualitative comparison of the design comparison, content-aware layout generation, and typography generation tasks, respectively.
From these results, we see that \approach{} significantly outperforms the baseline methods in design composition, and it even surpasses the specialized models in two design subtasks.

\newpage

\begin{coloredtextbox1}[A Data Example]{gray}
\textbf{HUMAN}: a poster of canvas width 1080px, canvas height 1920px. 
Please predict step by step according to the semantics of the elements.
After each prediction, there will be an intermediate rendering result as a reference to better make the next prediction.
Now predict the \textbf{background} elements: element 0: \texttt{<image>}

\textbf{ASSISTANT}: 
\begin{Verbatim}[frame=single]
{
    "index": 0,
    "left": 3,
    "top": -5,
    "width": 1101,
    "height": 460
}
\end{Verbatim}

\textbf{HUMAN}: current canvas state: \texttt{<image>}. 
Now predict the \textbf{underlay} elements: \texttt{null}

\textbf{ASSISTANT}: 
\begin{Verbatim}[frame=single]
{}
\end{Verbatim}

\textbf{HUMAN}: current canvas state: \texttt{<image>}. 
Now predict the \textbf{logo/image} elements: element 1: \texttt{<image>}

\textbf{ASSISTANT}: 
\begin{Verbatim}[frame=single]
{
    "index": 1,
    "left": -78,
    "top": 378,
    "width": 1228,
    "height": 1842
}
\end{Verbatim}

\textbf{HUMAN}: current canvas state: \texttt{<image>}. 
Now predict the \textbf{text} elements: element 2: \textit{Spring Clean}, element 3: \textit{Best hacks}

\textbf{ASSISTANT}: 
\begin{Verbatim}[frame=single]
{                                     {
    "index": 2,                           "index": 3,
    "left": 98,                           "left": 272,
    "top": 375,                           "top": 547,
    "width": 874,                         "width": 537,
    "height": 125,                        "height": 68,
    "angle": 0,                           "angle": 0,
    "font": "Raleway",                    "font": "Raleway",
    "font_size": 125,                     "font_size": 68,
    "color": [29, 29, 27],                "color": [0, 0, 0],
    "text_align": "center",               "text_align": "center",
    "capitalize": "false",                "capitalize": "false",
    "letter_spacing": 0.0,                "letter_spacing": 0.0,
    "line_height": 1.0,                   "line_height": 1.0
}                                     }
\end{Verbatim}

\textbf{HUMAN}: current canvas state: \texttt{<image>}. 
Now predict the \textbf{embellishment} elements: \texttt{null}

\textbf{ASSISTANT}: 
\begin{Verbatim}[frame=single]
{}
\end{Verbatim}

\end{coloredtextbox1}

\clearpage

\begin{figure*}[htbp]
    \centering
    \includegraphics[width=0.95\textwidth]{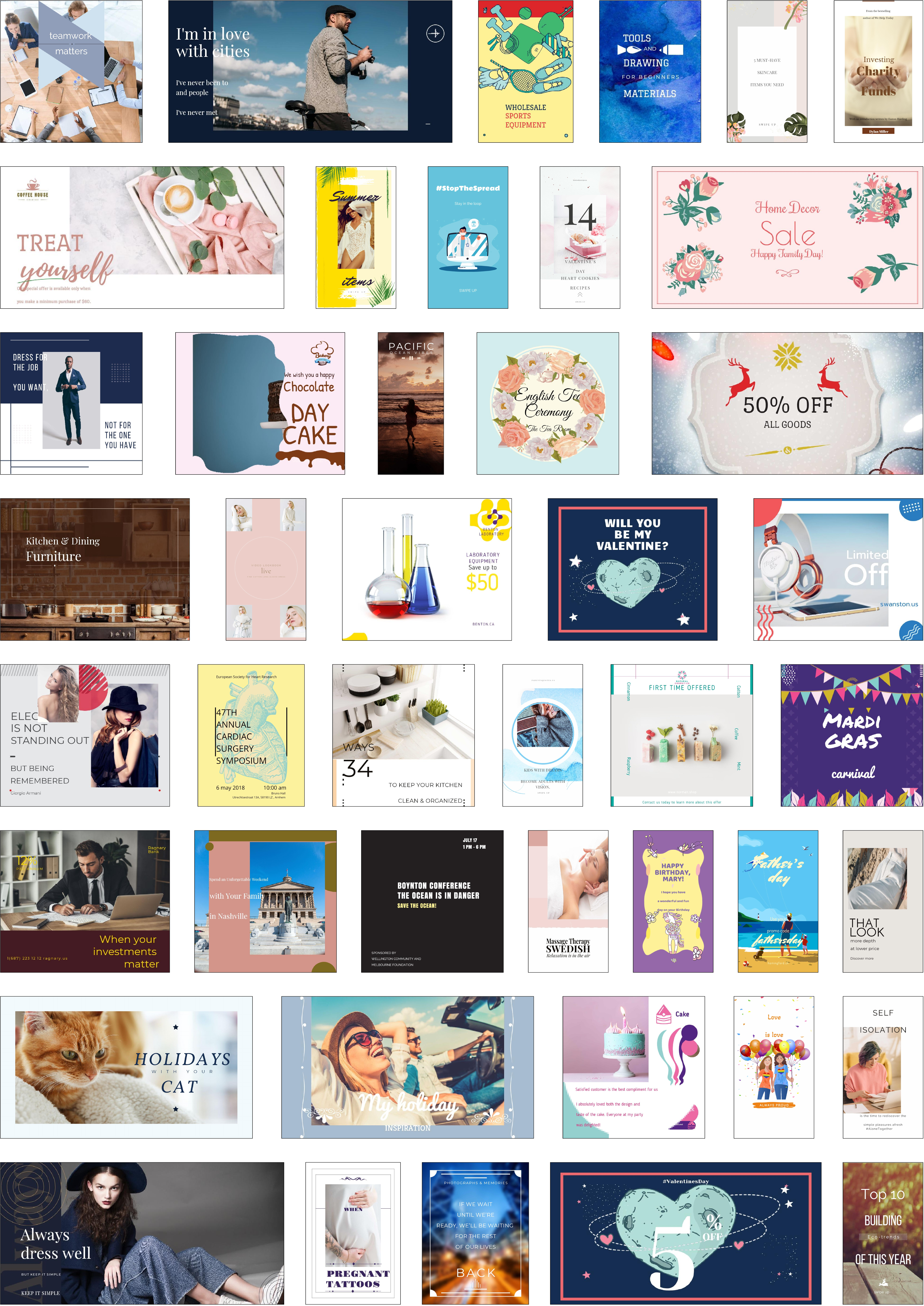}
\end{figure*}

\begin{figure*}[htbp]
    \centering
    \includegraphics[width=0.95\textwidth]{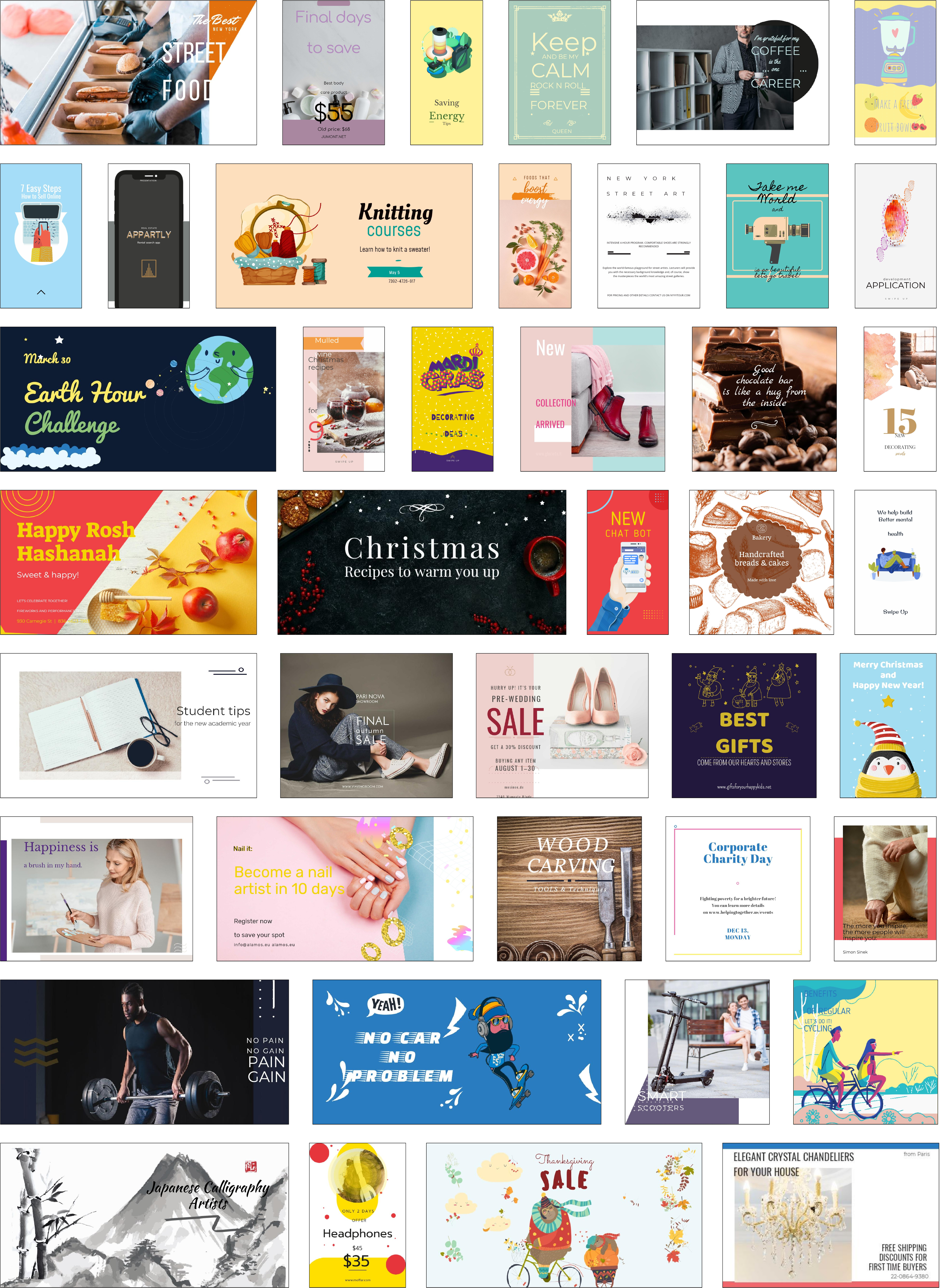}
\end{figure*}

\begin{figure*}[htbp]
    \centering
    \includegraphics[width=0.85\textwidth]{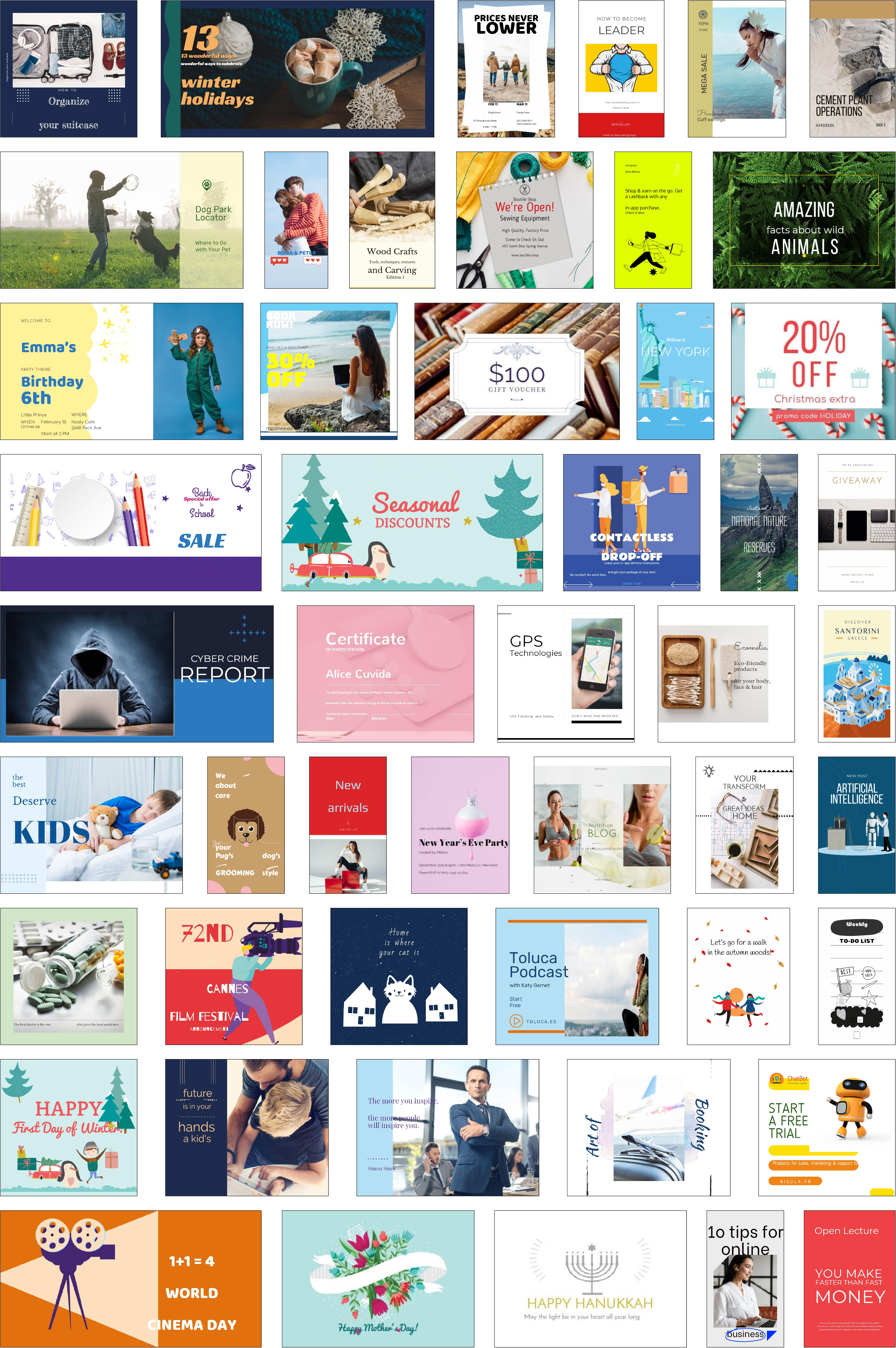}
    \label{fig:gallery}
    \caption{A gallery of graphic designs created by \approach{}.}
\end{figure*}

\begin{figure*}[htbp]
    \centering
    \includegraphics[width=0.75\textwidth]{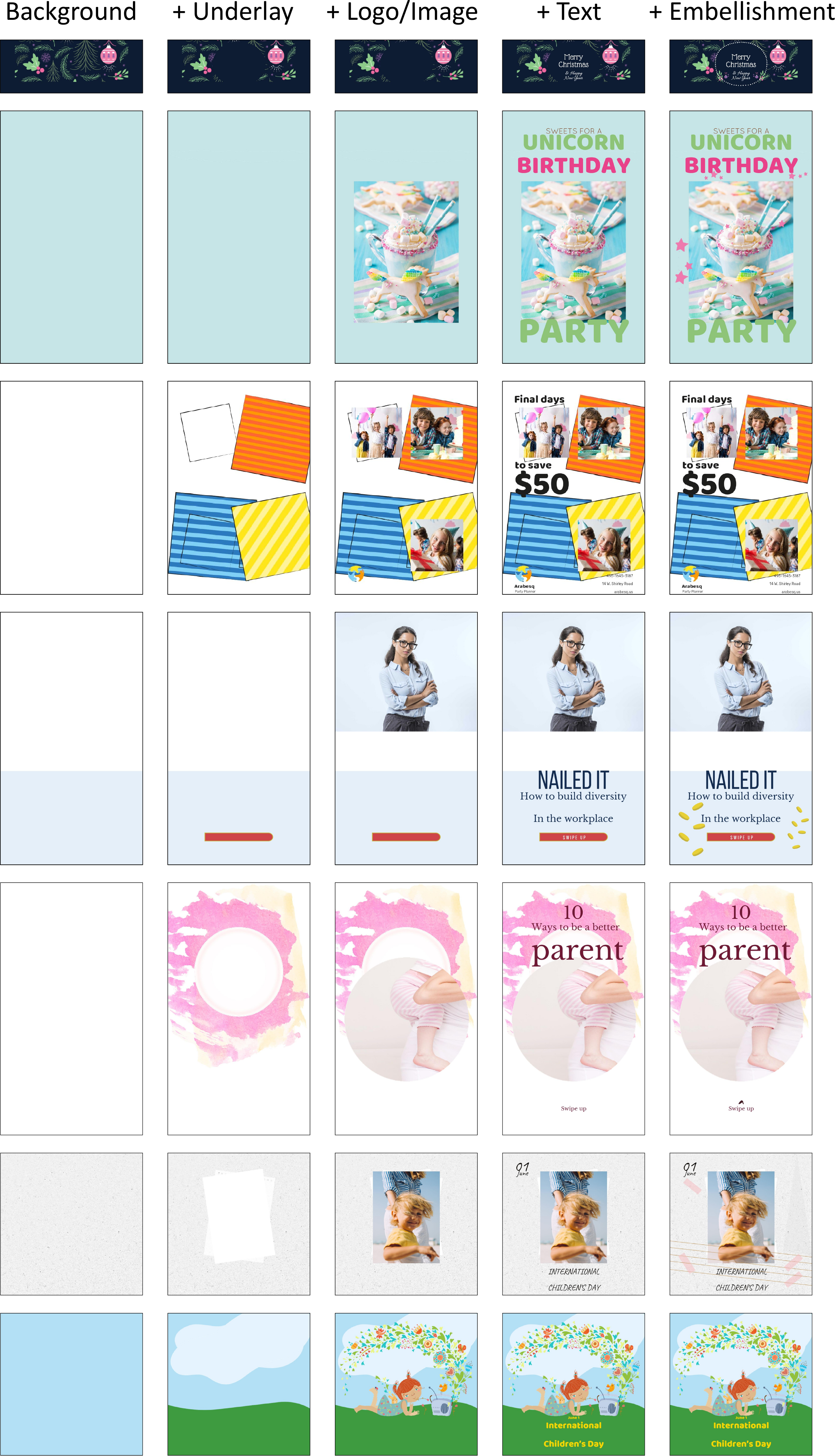}
    \label{fig:supp_layer}
    \caption{The layered design composition process in \approach{}. Our approach generates a holistic design in the order of background, underlay, logo/image, text, and embellishment layers.}
\end{figure*}

\begin{figure*}[htbp]
    \centering
    \includegraphics[width=0.75\textwidth]{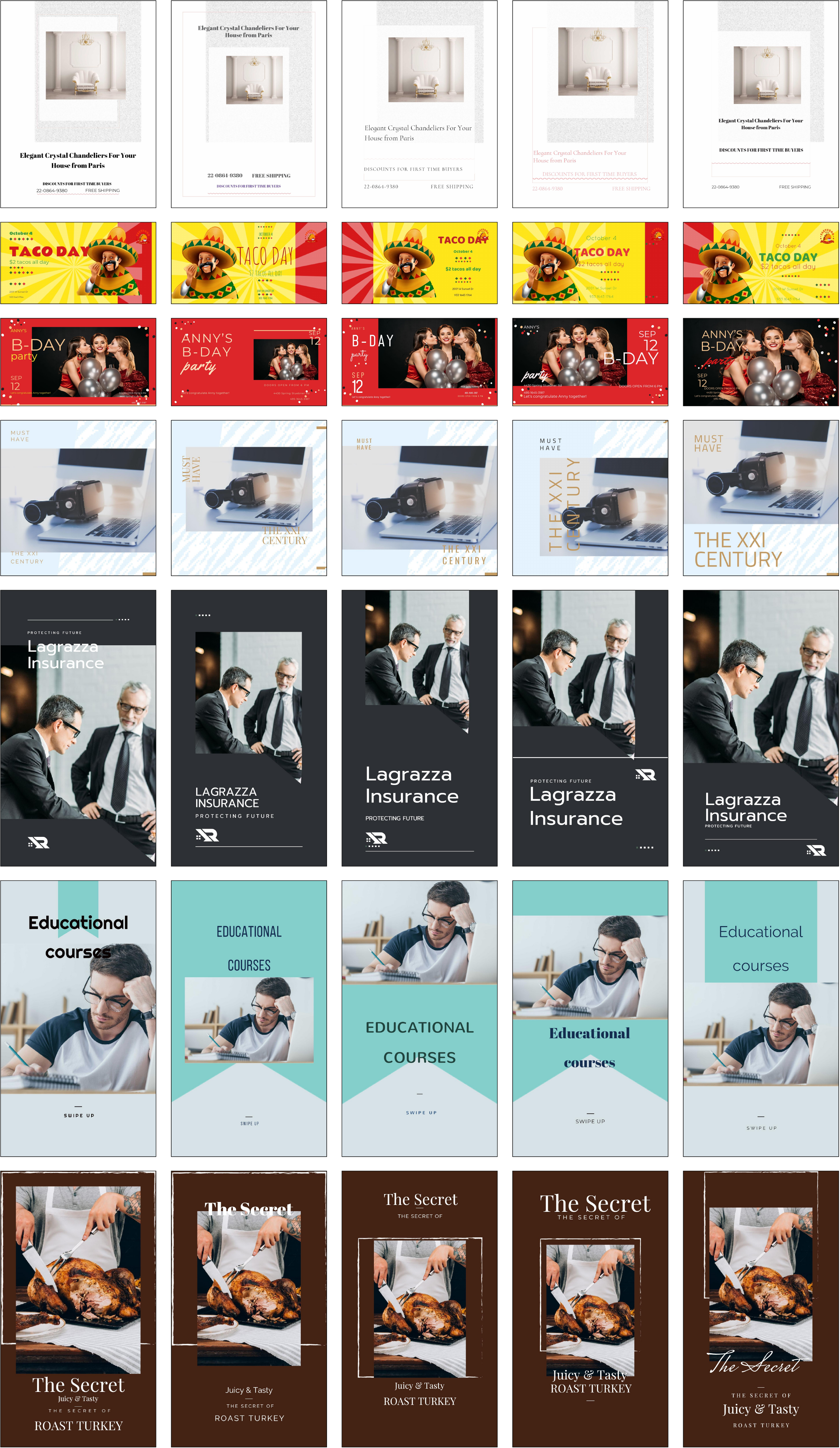}
    \label{supp_diversity}
    \caption{More results to demonstrate that \approach{} can create diverse designs with the same input.}
\end{figure*}

\begin{figure*}[htbp]
    \centering
    \includegraphics[width=0.95\textwidth]{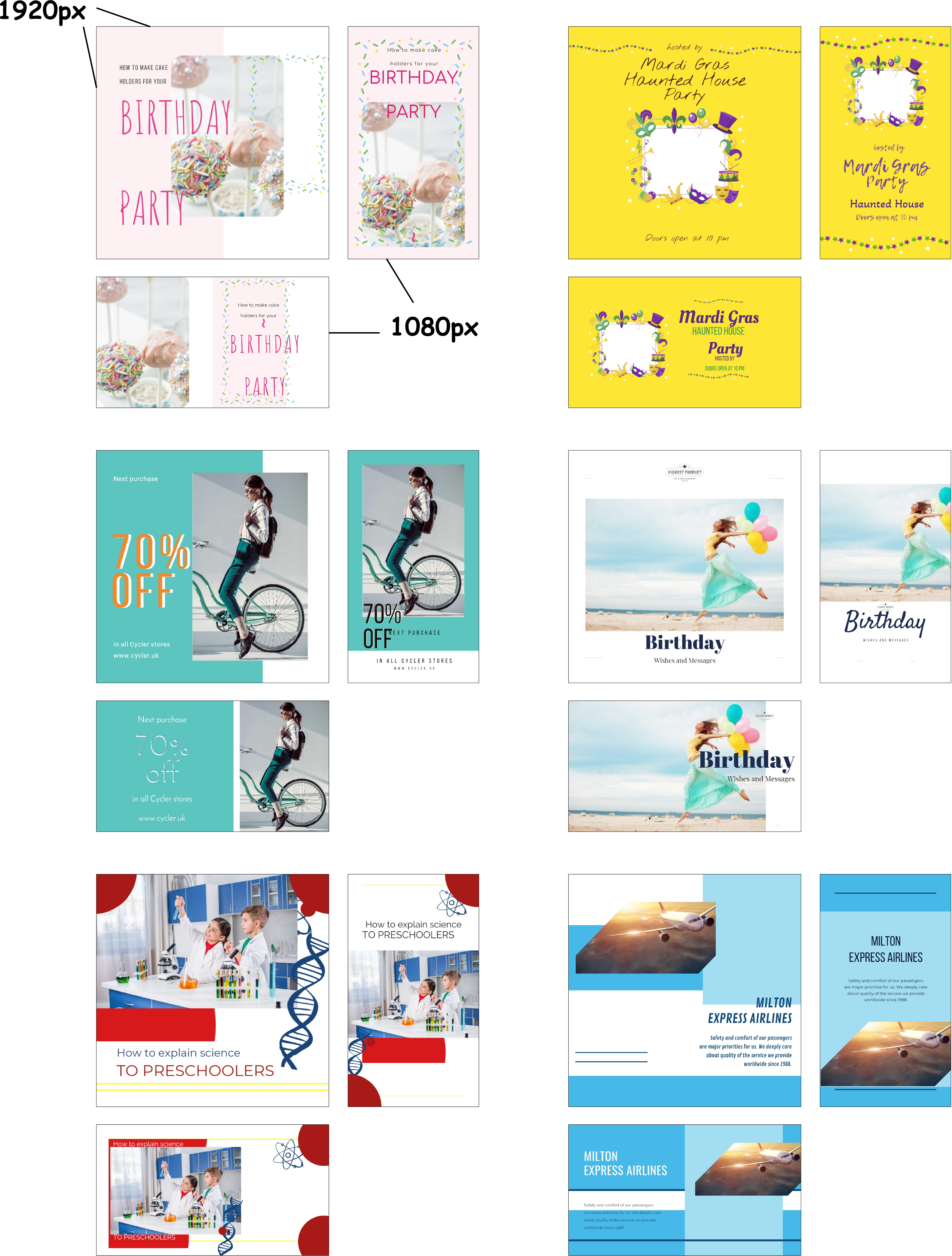}
    \label{supp_resolution}
    \caption{More results to demonstrate that \approach{} is able to generate graphic designs with different aspect ratios. }
\end{figure*}

\begin{figure*}[htbp]
    \centering
    \includegraphics[width=0.7\textwidth]{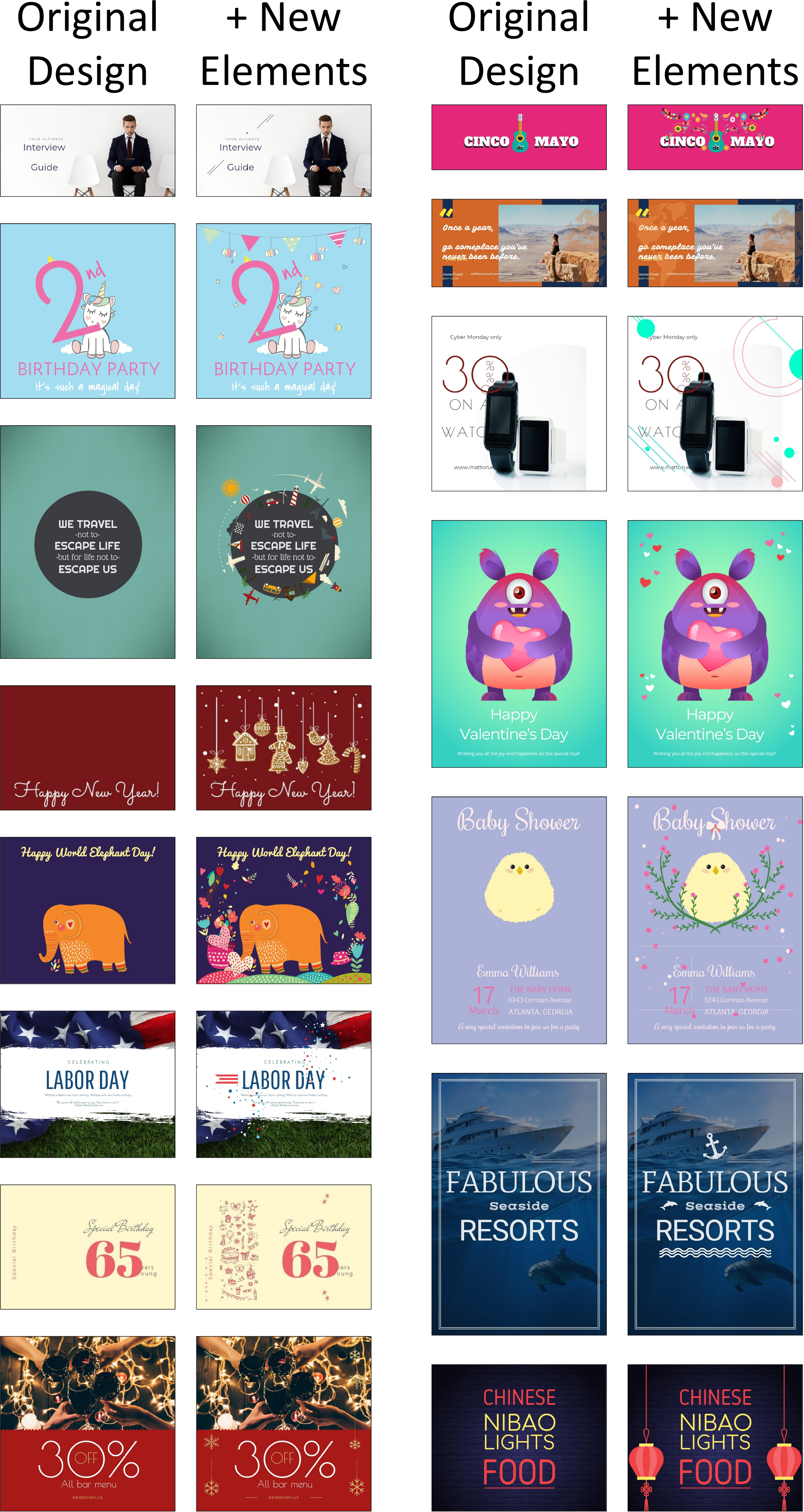}
    \label{supp_ele_filling}
    \caption{More results to demonstrate that \approach{} can add new elements to an existing design in a plausible way.}
\end{figure*}

\begin{figure*}[htbp]
    \centering
    \includegraphics[width=0.7\textwidth]{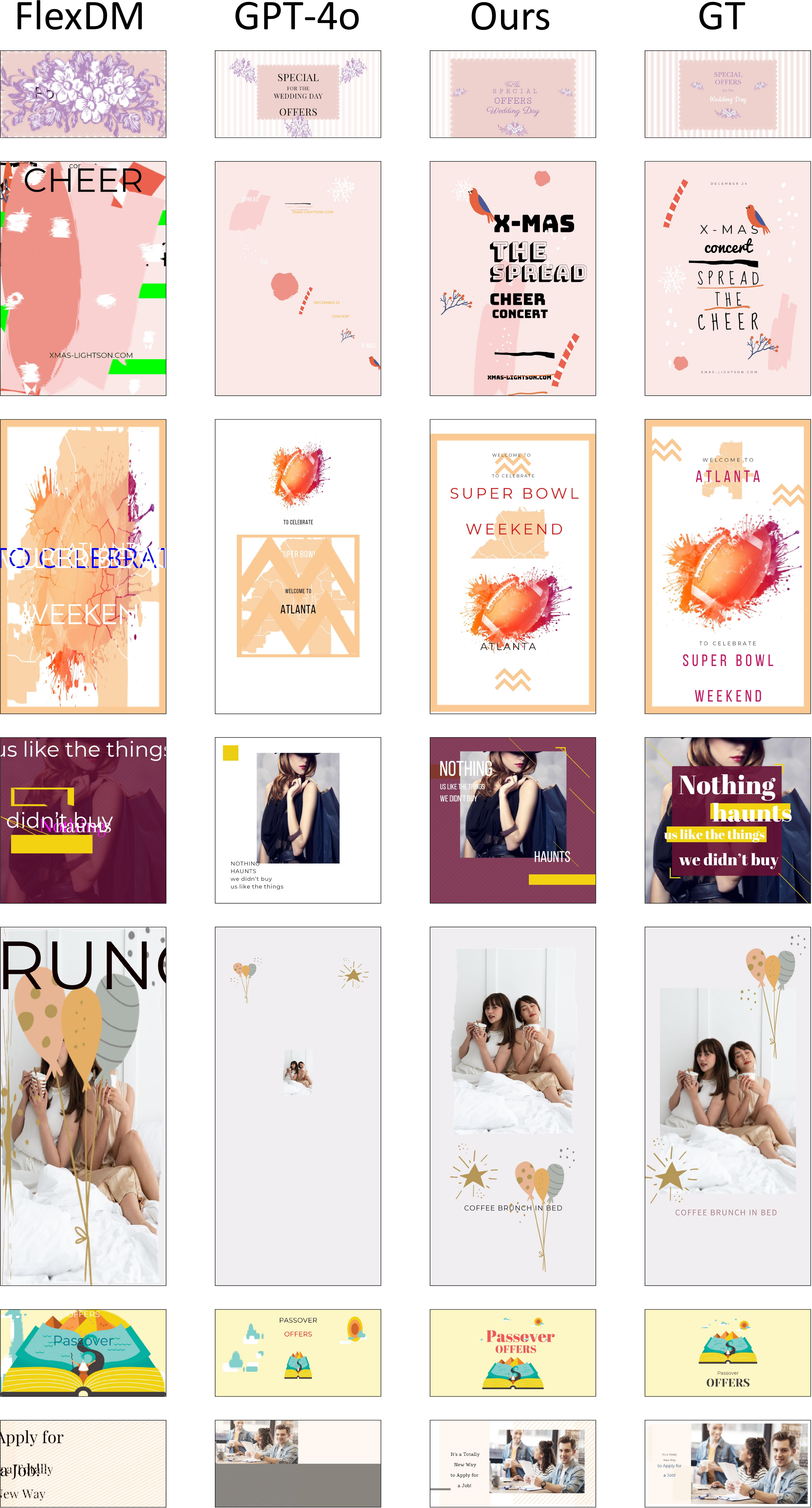}
    \label{supp_compare_overall}
    \caption{More qualitative comparison to demonstrate the superiority of \approach{} in design composition.}
\end{figure*}

\begin{figure}[htbp]
    \centering
    \includegraphics[width=0.5\textwidth]{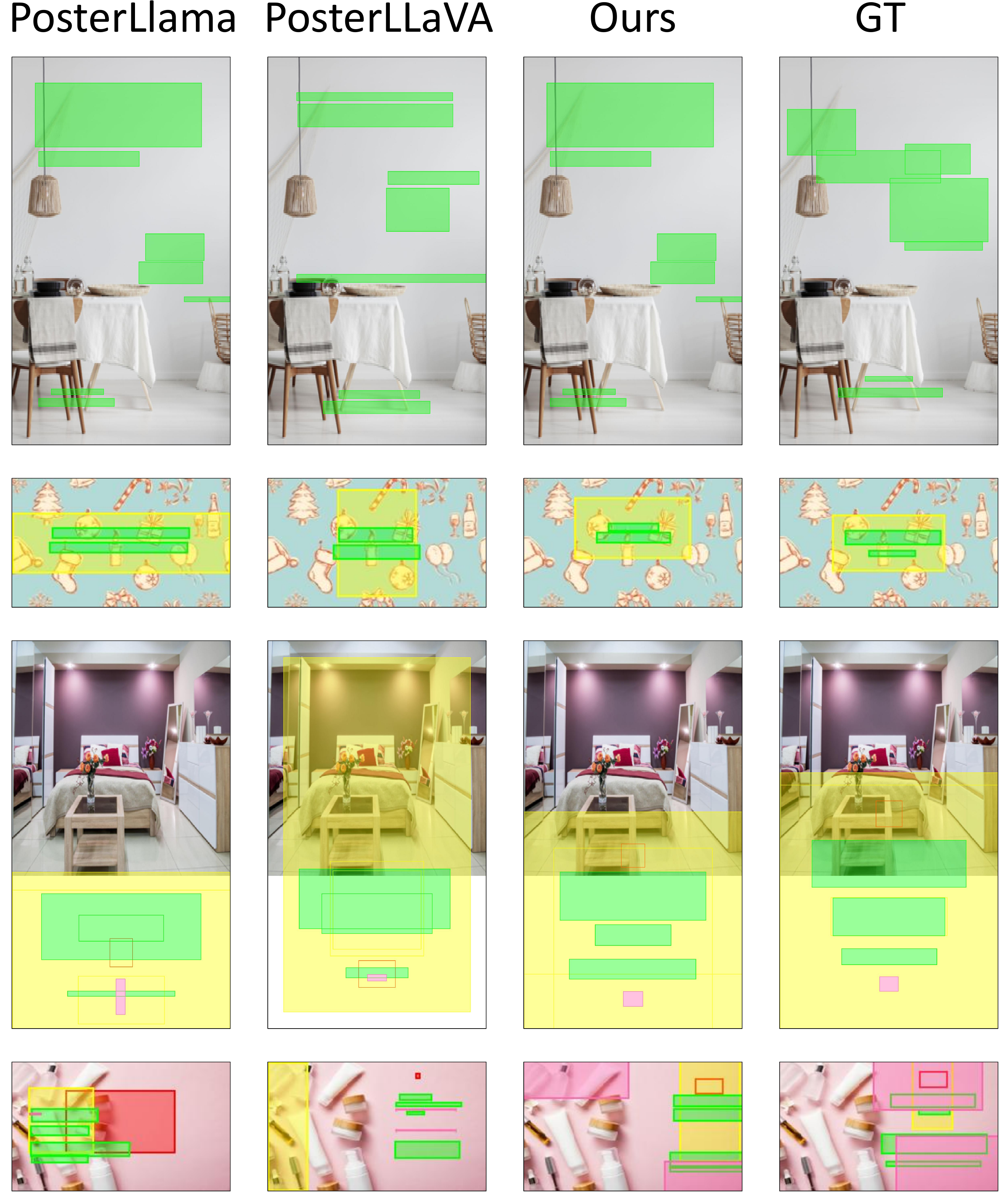}
    \label{supp_compare_content}
    \caption{More qualitative comparison to demonstrate the superiority of \approach{} in content-aware layout generation.}
\end{figure}

\begin{figure}[htbp]
    \centering
    \includegraphics[width=0.5\textwidth]{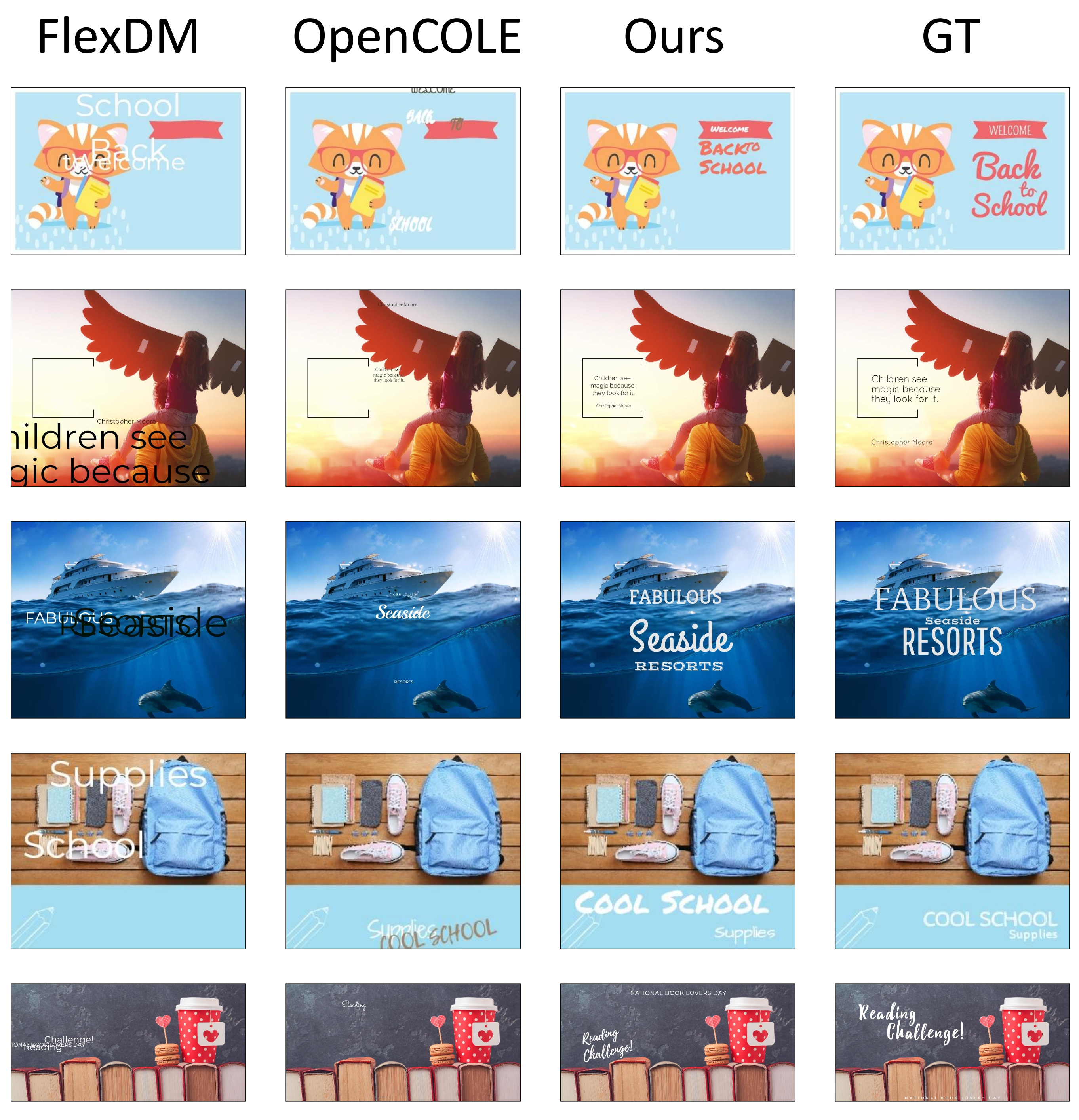}
    \label{supp_compare_typo}
    \caption{More qualitative comparison to demonstrate the superiority of \approach{} in typography generation.}
\end{figure}

